\documentclass[journal]{IEEEtran}

\usepackage{cite}
\usepackage{amsmath,amssymb,amsfonts}
\usepackage{algorithm}
\usepackage{algpseudocode}
\usepackage{graphicx}
\usepackage{textcomp}
\usepackage{bm}
\usepackage{dsfont}

\usepackage{subfig}
\usepackage{algorithm}
\usepackage{multirow}
\usepackage[dvipsnames]{xcolor}
\usepackage{url} 
\begin{document}

\title{On the Tradeoffs of On-Device Generative Models in Federated Predictive Maintenance Systems }

\author{Usevalad~Milasheuski,~\IEEEmembership{Graduate Student Member,~IEEE,}
        Piero~Baraldi,~\IEEEmembership{Senior Member,~IEEE,}
        Enrico~Zio,~\IEEEmembership{Fellow,~IEEE,}
        Stefano~Savazzi,~\IEEEmembership{Senior Member,~IEEE}
        % <-this % stops a space

\thanks{
%Manuscript received ...; revised ...; accepted ... . Date of current version ... .

This article is an extended version of our conference paper, ``Federated Generative Models for Predictive Maintenance in Industrial Environments,'' published in the \emph{Proceedings of the 2024 8th International Conference on System Reliability and Safety (ICSRS)}, 2024, pp.~156--161.

Usevalad Milasheuski is with Consiglio Nazionale delle Ricerche, IEIIT institute, 20133 Milan, Italy and also with the Dipartimento di Elettronica, Informazione e Bioingegneria (DEIB), Politecnico di Milano, 20133 Milan, Italy, (email: usevaladmilasheuski@cnr.it).

Piero Baraldi is with the Department of Energy, Politecnico di Milano, 20156 Milan, Italy (email: piero.baraldi@polimi.it).

Enrico Zio is with the Department of Energy, Politecnico di Milano, 20156 Milan, Italy and also Mines Paris - PSL, CRC, Sophia Antipolis, France (email: enrico.zio@polimi.it)

Stefano Savazzi is with Consiglio Nazionale delle Ricerche, IEIIT institute, 20133 Milan, Italy (email: stefano.savazzi@cnr.it).

The code is available at \protect\url{https://github.com/Bulbatronik/iot-journal}.}
}

% The paper headers
\markboth{Journal of \LaTeX\ Class Files,~Vol.~14, No.~8, August~2015}%
{Shell \MakeLowercase{\textit{et al.}}: Bare Demo of IEEEtran.cls for IEEE Journals}

\maketitle

\begin{abstract}
Federated Learning (FL) has emerged as a promising paradigm for preserving client data ownership and control over distributed Internet of Things (IoT) environments. While discriminative models dominate most FL use cases, recent advances in generative models — such as Variational Autoencoders (VAE), Generative Adversarial Networks (GAN), and Diffusion Models (DM) — offer new opportunities for unsupervised anomaly detection in time series analysis, with relevant applications in predictive maintenance (PdM) in critical industrial infrastructures. In this work, we present a comprehensive analysis of VAEs, GANs, and DMs in the context of federated PdM. We analyze their performance and communication overhead under both full and partial federation setups, where only subsets of model components are shared. Building on split-parameter personalized FL, this paper provides an empirical instantiation of shared/private parameter partitioning for unsupervised generative Time-Series Anomaly Detection (TSAD), mapping the split to \textit{analysis} and \textit{synthesis} modules across the generative architectures. Our experiments over five clients and two industrial time-series benchmarks reveal distinct trade-offs in model utility, stability, and communication efficiency in heterogeneous cross-silo FL. For the evaluated GAN-based configurations, full federation improves final anomaly-detection utility relative to independent local training, although these models remain less robust than the VAE- and DDPM-based alternatives.For DMs, decoder sharing matches full federation while substantially reducing the communication payload.

\end{abstract}


\IEEEpeerreviewmaketitle

% PART 1 Introduction
\section{Introduction} \label{sec:1_intro}

\IEEEPARstart{P}redictive maintenance (PdM) has emerged as a cornerstone of the Industrial Internet of Things (IIoT), playing a key role in the transition toward the next generation of industry (Industry 5.0) \cite{PDM1, PDM2}. By leveraging data-driven insights to anticipate equipment failures, PdM shifts operations from reactive or preventive maintenance to a proactive, condition-based approach. This paradigm aims to reduce costs and improve safety by preventing catastrophic system failures.

% Talk about predictive maintenance
The core enabling technology behind these PdM systems is Anomaly Detection (AD). Broadly defined as the identification of patterns that deviate from expected behavior, AD in time series (TS) is increasingly important across domains such as manufacturing \cite{AD_MANUFACT}, cybersecurity \cite{AD_CYBERSEC}, and healthcare \cite{AD_HEALTH}. By monitoring parameters such as temperature, vibration, and pressure, AD algorithms help PdM systems detect deviations from normal operating conditions, enabling timely interventions, reducing downtime, and extending machinery lifespan.

% Talk about Generative models
The effectiveness of AD depends on models capable of identifying anomalous patterns in large TS streams. Traditional methods, such as statistical tests, autoregressive models, and density-based techniques, often struggle to capture the complex temporal dependencies and multivariate interactions typical of streaming sensor data \cite{STREAM_DATA}. Deep Learning (DL) models have therefore been widely adopted for PdM. For example, recurrent neural networks (RNN), long short-term memory (LSTM), and convolutional neural networks (CNN) have been used in various time-series anomaly detection (TSAD) scenarios. Although DL methods improve on classical approaches by learning hierarchical temporal representations, they often fail to capture the full data distribution, especially under non-stationary or evolving conditions.

\begin{figure*}
\centerline{\includegraphics[width=0.99\textwidth]{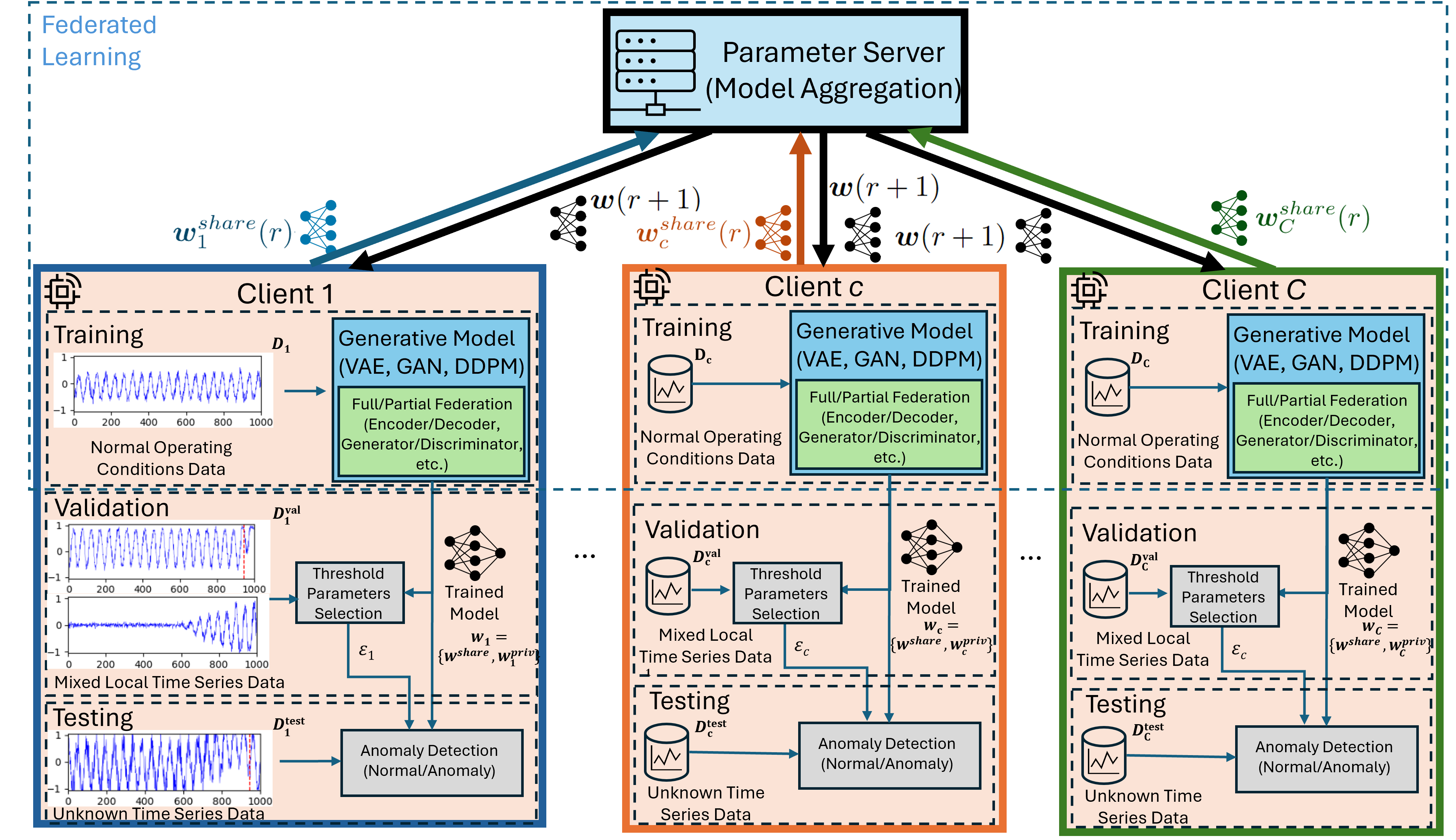} }
    \caption{Problem Setup for PdM TSAD. During the \textbf{training} stage, clients train the local model and transmit shared weight $\mathbf{w}_{c}^s$ to the server for multiple rounds. The \textbf{validation} stage individually estimates personalized anomaly threshold $\epsilon_{c}$ for each client. Finally, during \textbf{testing}, the clients evaluate the performance on the new time series data. }
\label{fig:system}
\end{figure*}

Recently, generative models emerged as powerful tools for unsupervised AD. These models learn sensor data under normal operating conditions and flag deviations through high anomaly score or low likelihood. Compared with discriminative models, this approach can better map realistic sensor patterns, augment limited datasets, and provide richer probabilistic insights \cite{GM_overview}. However, deploying generative AD models in industrial settings must address three main challenges:
\begin{itemize}
    \item \textbf{Data ownership and confidentiality}. Sensor streams often reveal sensitive operational details, and centralizing raw data from multiple plants may be infeasible because of IP or compliance constraints \cite{GDPR}.

    \item \textbf{Heterogeneous and non-IID distributions}. Different machines or plants often follow different dynamics (e.g., age, operating environment, load, wear), resulting in non-identically and independently distributed (non-IID) data \cite{HETEROGEN1}. As a result, a centrally trained model may perform poorly on local data or dilute localized anomalies when aggregating parameters from divergent distributions.

    \item \textbf{Resource, bandwidth, and energy constraints}. IoT edge devices and local gateways often have limited computation, energy, storage, and  bandwidth, constraining the transfer of large models or frequent updates \cite{Efff_IOT}.
\end{itemize}

These challenges motivate the integration of generative AD with ownership-preserving data management and communication-efficient training paradigms — most notably, \textit{Federated Learning (FL)} \cite{FEDAVG, FL_IoT_Survey}. While FL has been studied widely for classification or regression tasks, its adaptation to generative models - and in particular, for time series AD - remains largely unexplored \cite{FedSWTSAD, PARFED_MYSELF}. This gap highlights the need for a systematic evaluation of generative modeling in federated settings tailored to industrial IoT and PdM scenarios. Such an evaluation should consider not only detection performance, but also training stability, communication overhead, personalization, and robustness to heterogeneity.

\subsection{Related works}
The paradigm of AD has shifted from traditional unsupervised or self-supervised statistical boundary methods, such as One-class Support Vector Machines (OC-SVM) and Isolation Forest (IF), toward deep generative models \cite{ANOM_DET}. These models learn the probability distribution of normal data $P_{data}(\mathbf{x})$ \cite{GEN_ANOM_DET}, and identify anomalies either as samples in low-density regions of the learned distribution.

PdM has attracted considerable attention over the past decade because it is closely linked to TSAD \cite{PDM1, PDM4}. Its primary goal is to identify deviations from expected operational patterns, enabling early intervention and risk mitigation. By anticipating equipment failures before they occur, PdM shifts maintenance from reactive or preventive schedules to proactive, condition-based strategies. Through continuous monitoring of critical parameters such as temperature, vibration, and pressure \cite{PDM3}, PdM systems can detect subtle deviations, reduce downtime, extend machinery lifespan, and lower costs. In this context, AD seeks to model nominal system behavior in order to detect deviations. Generative modeling has emerged as an effective approach because it learns compact representations of input distributions and detects anomalies through reconstruction- or likelihood-based criteria. Among the most widely studied generative models for AD are Variational Autoencoders (VAEs) \cite{VAE}, Generative Adversarial Networks (GANs) \cite{GAN}, and Diffusion Models (DMs) \cite{DDPM}.

\textit{Variational Autoencoders (VAEs)} \cite{VAE} learn compressed representations that remain informative enough to reconstruct the input data. Several works have applied VAEs to industrial PdM AD. In \cite{VAE_AD_ADOPTED}, the authors replace the standard feed-forward layers of a VAE with LSTM units to capture temporal dependencies and sequential patterns in sensor data. The model in \cite{VAE_AD_2} enhances the standard sequential VAE by adding a smoothness-inducing term to the loss, encouraging smooth latent transitions. In contrast, \cite{VAE_AD_3} uses a standard VAE to compress short data windows into low-dimensional embeddings, and then trains an LSTM on these latent representations to predict future embeddings.

\textit{Generative Adversarial Networks (GANs)} \cite{GAN} originally gained popularity in image generation. In the context of time-series AD, works such as \cite{GAN_AD_ADOPTED, GAN_AD_1, GAN_AD_2} show promising adaptations of GANs to this problem. For example, MAD-GAN \cite{GAN_AD_1} is a straightforward GAN-based approach to AD, detecting anomalies through a combined score of reconstruction error and discrimination loss. Similarly, TAnoGAN \cite{GAN_AD_ADOPTED} uses LSTMs to capture temporal dependencies and combines a loss on hidden representations with the discriminator score for anomaly detection. During inference, for each test sequence, it runs an iterative optimization loop to find the latent vector that best reconstructs the sequence. Inspired by CycleGAN \cite{CycleGAN}, TadGAN \cite{GAN_AD_2} adopts a cycle-consistent architecture with two generators and two discriminators, ensuring that data can be mapped to the latent space and back, and yielding a more stable reconstruction error for anomaly scoring.

\textit{Diffusion Models (DMs)} \cite{DDPM} have emerged as an alternative to GANs for image generation, in some cases outperforming them \cite{DM_BEAT_GAN}. For AD, AnoDDPM \cite{DDPM_ADOPTED} departs from full denoising by introducing only partial corruption (e.g., $50\%$) and then reconstructing the original sample, thereby removing anomalies if present. The algorithm detects anomalies by comparing the reconstruction with the original input. TimeADDM \cite{DDPF_AD_1} employs a multi-step reconstruction strategy, corrupting data to different noise levels and attempting reconstruction from each of them. Anomalies are flagged when the model fails to recover the original signal from intermediate noise steps.

While generative models offer strong AD capabilities, they require large and diverse datasets that are rarely available at a single source. To address this limitation, \textit{Federated Learning (FL)} \cite{FEDAVG} offers a way to scale and accelerate PdM deployment across diverse critical equipment. By relying on \textit{model sharing} rather than \textit{data sharing}, FL lowers the administrative and legal burdens associated with information exchange. This is particularly important in industrial settings, where direct data exchange may be restricted by privacy regulations such as GDPR \cite{GDPR}. By aggregating local model updates, FL enables the global system to benefit from distributed datasets while preserving data ownership and reducing competitive risks.

Several studies have explored FL for generative models \cite{FEDGAN, FEDVAE, FEDDDPM}. However, it introduces substantial communication and computational overhead, especially for deep generative architectures with millions of parameters. This issue is particularly relevant in IoT-driven PdM, where edge devices are highly resource-constrained. To mitigate it, \textit{partial federation} has been proposed, in which only a subset of parameters is shared and jointly updated. Existing FL methods based on this idea have shown improved personalization \cite{FEDPER, FEDPER4}, greater robustness \cite{PARFED3, PARFED2}, and lower communication overhead \cite{PARFED_MYSELF}.

Despite these advances, FL of generative models in heterogeneous industrial IoT and PdM settings remains underexplored  \cite{FedSWTSAD, uFedHy-DisMTSADD,PrivTSAD-FedWGAN, FEDVAE}. Recent federated TSAD studies have begun to address related challenges through stronger AD pipelines. For example, FedSW-TSAD stabilizes adversarial training via Sobolev--Wasserstein constraints \cite{FedSWTSAD}, while PrivTSAD-FedWGAN adopts a federated WGAN-GP framework with a composite anomaly score for privacy-preserving multivariate TSAD \cite{PrivTSAD-FedWGAN}. Similarly, \cite{uFedHy-DisMTSADD} proposes a federated hypernetwork with client-specific transformer reconstruction models for distributed multivariate TSAD under heterogeneous settings. Beyond direct AD methods, recent federated generative tools such as FedTDD address misalignment through data distillation and synthetic-output exchange \cite{FEDTDD}. Although not an AD method, it remains relevant as a federated diffusion approach for heterogeneous TS data. Given the inherent challenges of training generative models, distributed learning over heterogeneous datasets can significantly degrade performance and hinder learning of the underlying data distribution \cite{HETEROGEN1}. This work aims to bridge this gap by systematically investigating federated model-partitioning strategies for generative AD in industrial IoT settings, substantially extending our preliminary VAE-based conference study \cite{PARFED_MYSELF}.

A related but broader research direction concerns the federated training of Time-Series Foundation Models (TSFMs). Recent works such as FFTS and FeDaL study federated pretraining and cross-dataset generalization for general-purpose TS representations, with AD treated as one downstream task among several others \cite{Chen2025FFTS,Chen2025FeDaL}. These developments are important to acknowledge, but they address a different setting from the one considered here: our focus is on on-device predictive maintenance with unsupervised AD from normal-condition data, client-specific threshold calibration, and explicit communication-efficiency trade-offs.

The general system model, illustrated in Fig. \ref{fig:system}, consists of distributed industrial clients that train local generative models on private sensor data and exchange model parameters with a server. The paper proposes a systematic partial-federation framework for private training of generative AD models in FL setups. Beyond performance evaluation, it aims to establish a taxonomy of generative methods for AD and to formulate guidelines for architecture selection for PdM in resource-constrained industrial IoT networks. We also investigate the performance and robustness of generative models under partial-federation policies designed to reduce communication overhead and mitigate instability, such as mode collapse, when aggregating heterogeneous generative models. Finally, we evaluate the framework for degradation-state assessment on multi-sensor systems operating in evolving environments.

A preliminary study \cite{PARFED_MYSELF} by the same authors introduced partial federation for PdM, comparing full, encoder-only, and decoder-only sharing against centralized and no-cooperation baselines. The present paper extends that contribution in scope, methodology, and evaluation. In particular, it moves from a single-model VAE study to a comparative analysis of VAE-, GAN-, and DDPM-based approaches, and reformulates partial federation through a unified \textit{analysis-versus-synthesis} perspective common to all model families and explicit communication overhead analysis. Accordingly, the paper analyzes how parameter-sharing policies affect unsupervised AD, personalization, and communication cost in federated PdM, rather than providing an exhaustive benchmark of optimizer-level personalized FL algorithms. Instead we study how partitioning \emph{generative} model into shared and private modules affects TSAD, personalization, stability, and communication cost. In this sense, the paper is structurally related to split-parameter personalized FL methods such as FedPer and FedRep, but differs in scope: rather than benchmarking optimizer-level personalization schemes, it develops and evaluates a taxonomy for unsupervised generative TSAD in federated PdM.

\subsection{Contributions}
The work explores the potential of FL and generative models to enable asset-level decentralized training of efficient AD tools for PdM. This paper studies a focused and practically relevant design question: how three representative families of unsupervised generative models for time-series anomaly detection (TSAD) behave in heterogeneous federated settings, and how partial sharing of their architectural components affects robustness, personalization, and communication efficiency. Rather than adopting a purely optimization-based view, we study FL through the lens of model partitioning: which parts of a generative architecture should be shared globally and which should remain client-specific under heterogeneous industrial conditions. In this sense, the proposed framework is structurally close to shared/private personalized FL based on model splitting, since only a subset of parameters is communicated and aggregated while the rest is retained locally \cite{FEDPER,FEDREP}. However, these algorithms are mainly tailored to supervised tasks such as classification, where the separation between representation learning and classifier components is straightforward. In unsupervised generative modeling, this separation is less clear because the task generally relies on input reconstruction. Accordingly, we treat the \textit{analysis/synthesis} decomposition as a practical design taxonomy and architecture-selection guideline for applying split-parameter personalized FL to unsupervised generative TSAD, rather than as a new FL paradigm. We specifically investigate the performance and robustness of generative models under several partial federation policies designed to reduce communication overhead and mitigate instability—such as mode collapse—often observed when aggregating heterogeneous generative models. We evaluate this framework for degradation-state assessment on multi-sensor systems operating in evolving environments.

The main contributions of this paper are as follows:
\begin{enumerate}
    \item We provide a structured review and problem framing of generative models for TSAD in privacy-preserving collaborative industrial settings, highlighting the current gap between recent advances in VAE-, GAN-, and DDPM-based anomaly detection and their still limited study under federated PdM setups.
    
    \item We instantiate split-parameter personalized FL for unsupervised generative TSAD by mapping the shared/private parameter split onto a common \textit{analysis/synthesis} decomposition of VAE-, GAN-, and DDPM-based models, so that full federation, partial federation, and fully local training are defined and compared consistently across model families.
    
    \item We develop a systematic partial-federation framework for unsupervised generative models under a common FL pipeline, so that the effect of shared/private architectural partitioning can be analyzed independently of optimizer-specific personalization mechanisms. This makes it possible to isolate how parameter-sharing policies influence anomaly-detection performance, personalization, stability, and robustness under heterogeneous federated PdM conditions.

    \item We provide an extensive empirical study that substantially extends our preliminary ICSRS 2024 conference paper. In particular, we move from a single-model VAE analysis to a comparative study of VAE-, GAN-, and DDPM-based approaches. We enrich the evaluation using different 
    micro- and macro-level criteria, including timing-aware PdM metrics to better separate the characteristics of each of the federation scenarios.

    \item We complement detection-performance results with a deployment-oriented analysis of federated PdM. Specifically, we quantify communication overhead in terms of transmitted parameters, exchanged traffic, and latency. Based on this joint analysis, the paper derives practical guidelines on when partial federation is preferable to full sharing or fully local training in resource-constrained industrial IoT environments.
\end{enumerate}

The paper is organized as follows. Section \ref{sec:gener_mod} introduces the generative models for AD, namely, the VAE, the GAN and DM-based approaches. In Section \ref{sec:fl}, we discuss the FL principles and their adaptation to personalization using partial federation. Section \ref{sec: ad_setup_models} provides the details of the problem setup and evaluation metrics. Section \ref{sect:case_study} provides the discussion of the experiment, including the dataset and models. Section \ref{sec:results} discusses the results of the simulations. Finally, Section \ref{sec: 5_concl} provides the conclusions of this work and future directions.
% PART 2 Related Works and Contributions
%\input{Sections/Sec_2_RelWorksAndContr}
% PART 3 Generative Modeling
\section{Overview of Generative Modeling} \label{sec:gener_mod}

\begin{figure*}[ht] \centerline{\includegraphics[width=0.85\textwidth]{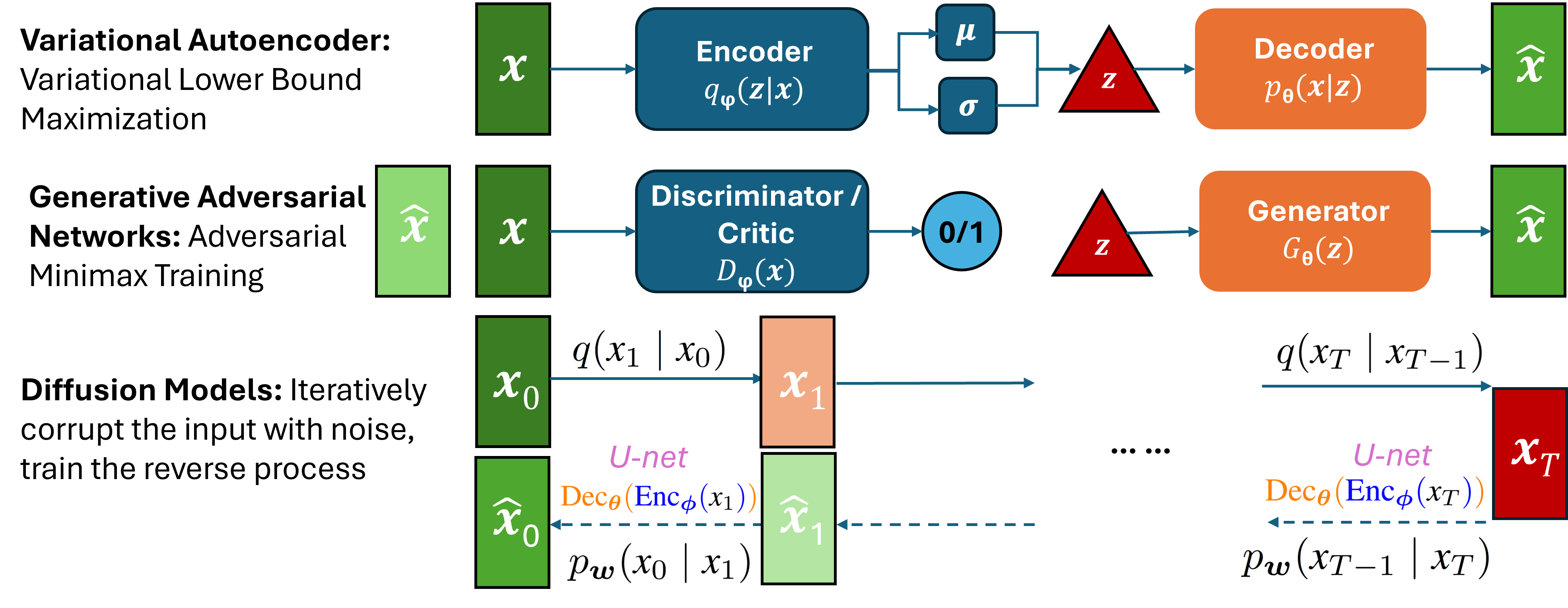} }
    \caption{A comparative overview of major generative model architectures: GANs, VAEs, and Diffusion Models.}
    \label{fig:gen_models}
\end{figure*}

In this section, we detail the three popular generative models suitable for deployment on edge IoT devices. As previously introduced, the core premise of unsupervised AD is learning the probability distribution of sensor time-series data collected during normal operating conditions. During the training stages, the model learns to reconstruct or generate samples within the manifold of normal behavior. During inference, significant reconstruction errors or low likelihood scores indicate anomalies \cite{GEN_ANOM_DET}. While discriminative models learn a decision boundary between normal (i.e., $y=0$) and abnormal (i.e., $y=1$) classes by modeling $P(y\mid\mathbf{x})$, generative models learn the underlying data distribution $P(\mathbf{x})$. In the context of AD where anomalous labels are scarce or non-existent, generative approaches are superior as they function in a purely unsupervised manner. They define normal behavior by what the model can generate, treating any deviation from this learned distribution as anomalous.
\subsection{Variational Autoencoders}
Variational Autoencoders (VAEs) \cite{VAE} are latent–variable generative models trained by maximizing a lower bound on the data log-likelihood. While using a similar idea to their Autoencoders (AEs) predecessors \cite{AE_AD_1}, it learns a continuous, well-structured latent distribution that enables principled generative sampling and improved generalization. The encoder $q_{\bm{\phi}}(\mathbf{z} \mid \mathbf{x})$ parametrized by $\bm{\phi}$ approximates the intractable posterior over latent variables $\mathbf{z}$, while the decoder $p_{\bm{\theta}}(\mathbf{x} \mid \mathbf{z})$ parametrized by $\bm{\theta}$ reconstructs data $\mathbf{x}$ from samples $\mathbf{z}$ drawn in the latent space. The encoder/decoder approach encourages the model to learn a meaningful structure within this latent space, which would be sufficient to approximate the input.

Training a VAE revolves around maximizing the Evidence Lower Bound (ELBO), which captures the trade-off between two complementary forces, namely the reconstruction error and the Kullback–Leibler (KL) divergence. The reconstruction term encourages the decoder to accurately reproduce the input data, while the KL divergence term gently regularizes the encoder’s distribution toward a simple prior, which is usually set to $p(\mathbf{z})=\mathcal{N}(0,I)$:
\begin{equation}
\mathcal{L}_{\text{ELBO}} =
\mathbb{E}_{\mathbf{z} \sim q_{\bm{\phi}}(\mathbf{z} \mid \mathbf{x})} \big[ \log p_{\bm{\theta}}(\mathbf{x} \mid \mathbf{z}) \big]
- \mathrm{KL} \big( q_{\bm{\phi}}(\mathbf{z} \mid \mathbf{x}) \,\|\, p(\mathbf{z}) \big).
\label{eq:vae_short_elbo}
\end{equation}

A crucial element enabling this optimization is the reparameterization trick. Instead of sampling $\mathbf{z}$ directly from the encoder’s distribution, the model expresses each latent sample as a deterministic transformation of noise, which is called \textit{reparameterization trick}:
\begin{equation}
\mathbf{z} = \mu_{\bm{\phi}}(\mathbf{x}) + \sigma_{\bm{\phi}}(\mathbf{x})\odot\epsilon, \quad \epsilon \sim \mathcal{N}(\mathbf{0}, \mathbf{I}).
\end{equation}
This reformulation allows gradients to pass cleanly through the sampling step, making end-to-end learning possible.

\subsection{Generative Adversarial Networks}
Generative Adversarial Networks (GANs) \cite{GAN} train a
generator $G_{\bm{\theta}}(\mathbf{z})$ and discriminator $D_{\bm{\phi}}(\mathbf{x})$ in a minimax setup where the discriminator $D_{\bm{\phi}}$ distinguishes real samples from generated ones and the generator $G_{\bm{\theta}}$ produces synthetic data in an attempt to fool the discriminator $D_{\bm{\phi}}$. The standard objective can be formulated in the following way

\begin{equation}
\min_{G_{\bm{\theta}}} \max_{D_{\bm{\phi}}} \;
\mathbb{E}_{\mathbf{x} \sim p(\mathbf{x})} \big[ \log D_{\bm{\phi}}(\mathbf{x}) \big]
+ \mathbb{E}_{\mathbf{z} \sim p(\mathbf{z})} \big[ \log \big( 1 - D_{\bm{\phi}}(G_{\bm{\theta}}(\mathbf{z})) \big) \big].
\label{eq:gan_loss_short2}
\end{equation}

Although effective, this formulation presents well-known challenges \cite{TrainingGAN}. In practice, adversarial min--max optimization is often unstable and requires careful balancing between the generator and discriminator during training. Problems such as vanishing gradients, sensitivity to hyperparameters, and oscillatory convergence are common, especially on complex or heterogeneous data distributions \cite{GAN_COLLAPSE}. Much of the progress in GAN research has therefore focused on stabilizing these dynamics. A key step in this direction is the \textit{Wasserstein GAN} \cite{WGAN}, which replaces the Jensen--Shannon divergence of the original GAN with the Wasserstein-1 distance. This change turns the discriminator into a \emph{critic} $f_{\bm{\phi}}$, removes the sigmoid, and allows real-valued outputs:
\begin{equation}\label{eq:wgan}
\max_f \;
\mathbb{E}_{\mathbf{x}}[f_{\bm{\phi}}(\mathbf{x})] - \mathbb{E}_{\mathbf{z}}[f_{\bm{\phi}}(G_{\bm{\theta}}(\mathbf{z}))].
\end{equation}
The generator then minimizes the same expression. This redefinition yields smoother gradients and training behavior but introduces a new requirement: the critic must satisfy a 1-Lipschitz constraint. The improved WGAN-GP \cite{WGAN_GP} introduces a soft constraint for Eq. \ref{eq:wgan} by penalizing deviations of the gradient norm from 1 on points interpolated between real and generated samples:
\begin{equation}
    \lambda \,\mathbb{E}_{\tilde{\mathbf{x}}}
    \left( \lVert \nabla_{\tilde{\mathbf{x}}} f_\mathbf{\phi}(\tilde{\mathbf{x}}) \rVert_2 - 1 \right)^2,
\end{equation}
where $\tilde{\mathbf{x}}$ is defined as a random interpolation between a real data sample $\mathbf{x}$ and a generated sample $\hat{\mathbf{x}}$. This gradient penalty ensures the critic satisfies the 1-Lipschitz continuity constraint along the straight lines connecting real and generated data, resulting in smoother gradients and a more stable training.

\subsection{Denoising Diffusion Probabilistic Models}

Denoising Diffusion Probabilistic Models (DDPMs) \cite{DDPM} take a fundamentally different approach to generative modeling by learning to reverse a gradual corruption of data. Instead of mapping noise to data in a single step, they construct generation as a careful walk backward through a noising process. At training time, a clean data sample $\mathbf{x}_0$ is progressively corrupted through a fixed Markov chain:
\begin{equation}
q(\mathbf{x}_t \mid \mathbf{x}_{t-1}) = \mathcal{N}\!\left(\sqrt{1-\beta_t}\, \mathbf{x}_{t-1}, \; \beta_t \mathbf{I} \right),
\end{equation}
where $\{\beta_t\}_{t=1}^T$ is a variance schedule. The closed-form expression
for $q(\mathbf{x}_t \mid \mathbf{x}_0)$ allows sampling noisy versions of $\mathbf{x}_0$ at arbitrary
timestep~$t$.

A model is trained to predict the noise added at each step. The most common architecture used for this purpose is U-net \cite{UNET}, which consists of an encoder–decoder structure with skip connections that preserve multi-scale information across resolutions. This design enables effective modeling of both local and global temporal dependencies in time-series data, making it well-suited for the iterative denoising process central to DDPMs. We define this model as $\epsilon_{\bm{w}}(\mathbf{x}_t, t)$ parametrized by $\bm{w}$, where $\bm{w}=\{\bm{\phi}, \bm{\theta}\}$ represent the parameters of the encoder and decoder parts of the U-net model, respectively. The variational bound leads to a simplified denoising
objective:
\begin{equation}
\mathcal{L}_{\text{DDPM}} = \mathbb{E}_{t,\,\mathbf{x}_0,\,\epsilon} \left[ \left\lVert \epsilon - \epsilon_{\bm{w}}(\mathbf{x}_t, t) \right\rVert_2^2 \right],
\end{equation}
where $x_t$ is produced via the forward noising process
$\mathbf{x}_t = \sqrt{\bar{\alpha}_t}\, \mathbf{x}_0 + \sqrt{1 - \bar{\alpha}_t}\, \epsilon$
and $\bar{\alpha}_t = \prod_{s=1}^t (1 - \beta_s)$.

Sampling proceeds by iteratively denoising from $\mathbf{x}_T \sim \mathcal{N}(\mathbf{0},\mathbf{I})$
back to $\mathbf{x}_0$ using learned mean predictions:
\begin{equation}
p_{\bm{w}}(\mathbf{x}_{t-1} \mid \mathbf{x}_t) = \mathcal{N}\!\left( \mu_{\bm{w}}(\mathbf{x}_t, t), \; \sigma_t^2 I
\right),
\end{equation}
yielding high-quality generative samples.

When it comes to AD, these models generally rely on thresholding over a particular score function. Generally, this function is a loss function, such as Mean Squared Error (MSE) or Mean Absolute Error (MAE) between the original input $\mathbf{x}$ and its reconstructed version of the input $\hat{\mathbf{x}}$, like for VAE in \cite{VAE_AD_ADOPTED} and DDPM in \cite{DDPM_ADOPTED}. However, comparison of the activations of the intermediate representations between the real and generated inputs, $h(\mathbf{x})$ and $h(\hat{\mathbf{x}})$, may provide a more meaningful anomaly signal, as in \cite{GAN_AD_ADOPTED}.

\subsection{Unified architectural perspective}

VAE-, GAN-, and DDPM-based models can be decomposed into two fundamental functional components:
(i) an \emph{analysis} component, which maps high-dimensional TS observations into latent, intermediate, or scored representations, and
(ii) a \emph{synthesis} (or generation) component, which reconstructs or generates data samples from such representations.

In VAEs, this decomposition naturally corresponds to the encoder--decoder pair. In GANs, the discriminator (or critic) fulfills the analysis role by extracting informative features from data samples and outputting a signal on whether a given sample comes from the true data distribution, while the generator acts as the synthesis module by mapping latent noise to the data space. Similarly, in DDPMs, the U-Net backbone can be interpreted as a structured encoder--decoder architecture, where the down-sampling path encodes noisy observations into hierarchical representations and the up-sampling path performs the denoising and reconstruction process.

This unified perspective allows all models to be described using a common parameter partition $\{\bm\phi, \bm\theta\}$, independently of their specific architectural choices. As clarified in the following section, this perspective forms the basis for the federation strategies across heterogeneous architectures. In other words, the proposed \emph{partial federation} strategies can be formulated by selectively sharing either the analysis component ($\bm\phi$), the synthesis component ($\bm\theta$), or both, while keeping the remaining parameters local.

% PART 4 Privacy-Preserving Machine Learning
\section{Federated System for Generative Model Training} \label{sec:fl}
To address the challenges of data privacy, regulatory compliance (e.g., GDPR), and bandwidth constraints in industrial IoT, we leverage FL. This section outlines the standard FL formulation and introduces the \textit{partial federation} paradigm, which is aimed at enhancing personalization and training stability in generative models while reducing communication overhead. Building on this framework, we discuss how \textit{partial federation} can be systematically applied to the VAEs, GANs, and DDPMs generative architectures illustrated before, highlighting model-specific federation strategies and shared component designs.
\subsection{Federated Learning and Data Heterogeneity}
FL enables the collaborative training of a global model across multiple decentralized edge devices (clients) without exchanging raw local data. Consider a set of $C$ industrial sites (clients), where each client $c$ possesses a private local dataset $\mathcal{D}_c$ of time-series sensor readings. The goal is to minimize a global objective function $\mathcal{L}(\bm{w})$ over the model parameters $\bm{w}$:

\begin{equation}    
    \min_{\bm{w}} \mathcal{L}(\bm{w}) = \sum_{c=1}^{C} \frac{|\mathcal{D}_c|}{|\mathcal{D}|}\mathcal{L} _c(\bm{w}),
\end{equation}
where $|\mathcal{D}| = \sum_c |\mathcal{D}_c|$ is the total data volume, and $ \mathcal{L}_c(\bm{w}) = \mathbb{E}_{\mathbf{x} \sim \mathcal{D}_c}[\mathcal{L}(\mathbf{x}, \bm{w})]$ is the local empirical risk on client $c$.

In the standard FedAvg \cite{FEDAVG} algorithm, the server initializes a global model $\bm{w}(0)$. At each communication round $r$:
\begin{enumerate}
 
    \item The server broadcasts the current global parameters $\bm{w}(r)$ to a subset of eligible clients $F(r)$.
    \item Each client $c$ updates the model locally using Stochastic Gradient Descent (SGD) on its private data $\mathcal{D}_c$, producing an updated local model $\bm{w}_c(r+1)$.
    \item The server aggregates these local updates to produce the next global model $\bm{w}(r+1)$.
\end{enumerate}

While effective for standard supervised tasks, FedAvg faces significant hurdles in generative AD for IoT: 
\begin{itemize}
    \item \textbf{non-IID heterogeneous data}: Industrial assets often operate under distinct environmental conditions or load profiles (heterogeneity). Averaging generative models (like VAEs or GANs) trained on divergent distributions often leads to "mode averaging," where the global model fails to capture the sharp, specific distributions required for accurate reconstruction.
    \item \textbf{Communication Overhead}: Deep generative models (e.g., Diffusion Models) may contain millions of parameters. Transmitting full gradients or weight updates frequently over bandwidth-constrained IoT networks is often infeasible.
\end{itemize}

\subsection{Partial Federation Paradigm}
To mitigate the limitations of full federation, we adopt a \textit{partial federation} strategy. In this setup, the model parameters $\bm{w}$ are partitioned into two disjoint sets: shared parameters $\bm{w}^{share}$, which are communicated between each client and the server, and private parameters $\bm{w}_c^{priv}$, which remain local throughout the federation process. Thus, we define the global model $\bm{w}(r) = \{\bm{w}(r)\}$, while the local model is defined as $\bm{w}_c(r) = \bm{w}(r) \cup\bm{w}_c^{priv}(r)$. In the classic FedAvg case, our algorithm reduces to $\bm{w}^{share}=\{\bm{w}\}$ and $\bm{w}_c^{priv}(r)= \{ \emptyset\}$. In contrast,  $\bm{w}^{share}=\{\emptyset \}$ and $\bm{w}_c^{priv}(r)= \{\bm{w} \}$ correspond to independent local training without communication. Whenever both, $\bm{w}_c^{priv}(r)\neq \{ \emptyset\}$ and $\bm{w}_c^{share}(r)\neq \{ \emptyset\}$, the approach can be interpreted as adaptation of personalized FL algorithms, such as FedPer \cite{FEDPER} or FedRep \cite{FEDREP} for unsupervised AD.

More details on the training procedure are provided in Algorithm~\ref{alg:fedavg}. This modular approach allows operators to balance global knowledge discovery, local anomaly sensitivity, and personalization \cite{PARFED_MYSELF}. Within this formulation, the paper deliberately focuses on the effect of model partitioning itself.

In the preceding section, we categorized the three main generative frameworks: VAEs, GANs, and DDPMs. These architectures allow for a strategic partitioning of components based on their functional roles. In VAEs, the encoder ($\bm{\phi}$) performs amortized inference by compressing input data into a low-dimensional latent space, while the decoder ($\bm{\theta}$) reconstructs the original data manifold. Similarly, in the GAN framework, the discriminator (or critic, $\bm{\phi}$) acts as a feature extractor that maps data to an informative scalar signal (the probability of being real), while the generator ($\bm{\theta}$) acts as the generative counterpart, transforming noise into synthetic samples. In DDPMs, the U-Net backbone \cite{UNET} is structurally split: the encoder portion ($\bm{\phi}$) progressively downsamples the input to capture high-level hierarchical features, whereas the decoder ($\bm{\theta}$) upsamples these features to reconstruct the denoised output (Fig. \ref{fig:gen_models}). By unifying these components under the notations $\bm{\phi}$ and $\bm{\theta}$, we highlight a fundamental duality: $\bm{\phi}$ (which we refer to as the \textit{encoder}) parameters are associated with the analysis of data (mapping from high-dimensional input to compressed representations or scores), whereas $\bm{\theta}$ (which we refer to as the \textit{decoder}) parameters are dedicated to data synthesis (mapping from latent or noisy representations back to the data space). This gives flexibility in defining $\bm{w}^{share}$ and $\bm{w}^{priv}$ by selecting whether to share/store $\bm{\phi}$, $\bm{\theta}$ or $ \{\bm{\phi}, \bm{\theta}\}$. In this sense, the proposed formulation is related to split-parameter personalized FL methods such as FedPer \cite{FEDPER} and FedRep \cite{FEDREP}, which also distinguish shared and client-specific parameters. However, those methods target supervised models, where splitting a shared representation from a local prediction head is natural. Here, we consider unsupervised generative TSAD, where the relevant partition instead lies between analysis and synthesis modules that jointly affect reconstruction quality, anomaly sensitivity, and training stability. Rather than benchmarking optimizer-level personalization schemes, we focus on the architectural question of which generative components should be shared or kept local under heterogeneous industrial conditions.

When the analysis module is shared, clients are encouraged to map local observations into a more aligned latent representation while retaining client-specific reconstruction through private synthesis modules. Conversely, when the synthesis module is shared, clients preserve local feature extraction but reconstruct through a common generative pathway. Under heterogeneous industrial conditions, these two choices encode different trade-offs between global alignment and local specialization. This rationale motivates the empirical comparison developed in the rest of the paper.

Optimizer-level personalized FL methods such as FedProx~\cite{FedProx}, SCAFFOLD~\cite{SCAFFOLD}, MOON~\cite{MOON} are complementary to the architectural partitioning studied here. However, in this work, we keep the federation optimizer fixed to isolate the effect of shared/private parameter placement in generative TSAD models, while combining both directions is left as future work.

\begin{algorithm}[t] % FedAvg 
\caption{Partial Federation with generative models}
\begin{algorithmic}[1]
\State \textbf{Input:} Global parameters $\bm{w}(0)=\{ \bm{w}^{share}(0) \}$, number of clients $C$, private parameters $\bm{w}_c^{priv}(0)$, number of communication rounds $R$

\For{each round $r = 1, 2, \ldots, R$}
  \State Server selects a subset $F(r)$ of $C$ clients
  \For{each client $c \in F_r$ \textbf{in parallel}}
    \State Client $c$ receives $\bm{w}(r)$ from the server
      \State $\bm{w}_c(r) \leftarrow \{\bm{w}(r) \cup \bm{w}_c^{priv}(r)\}$
  \For{each local epoch $i$ from $1$ to $E$}
    \For{each batch $b$ of data from $D_c$}
      \State $\bm{w}_c(r) \leftarrow \bm{w}_c(r) - \eta \nabla \ell(\bm{w}_c(r); b)$ 
    \EndFor
  \EndFor
    \State Client $c$ sends  $\bm{w}_c^{share}(r)$ of $\bm{w}_c(r)$ to the server
  \EndFor
  \State Server aggregates the updates:
    \State $\bm{w}(r+1) \leftarrow \sum_{c \in F_r} \frac{|D_c|}{|D_r|} \bm{w}_c^{share}(r)$, where $|D_c|$ is the number of data points on client $c$ and $|D_r| = \sum_{c \in F_r} |D_c|$
\EndFor
%\State \textbf{Output:} Trained global model $\bm{w}(T)$
\end{algorithmic}\label{alg:fedavg}
\end{algorithm}

% PART 5 Models Setup
\section{Distributed Anomaly Detection and Models Setup}\label{sec: ad_setup_models}
 In this section, we discuss the problem and the distributed AD system and the deployed models. We briefly explain how these models detect the anomaly and discuss the performance metrics used for our experiments.
%\vspace{-0.7cm}
\subsection{Distributed Anomaly Detection}\label{subsec:dataset}
For our problem, we consider a set of $C$ clients, each equipped with a health monitoring system of $K$ sensors. The sensor measurements obtained from different clients $c$ share different factors, such as geolocation and operating conditions. The operating machine is designed to operate for $T$ arbitrary time units (atu) unless a random failure occurs. Each measurement sensor operates at a fixed frequency $f_s = 1$ $\text{atu}^{-1}$. We assume each of the $C$ clients has a stored history of the equipment runs of $|D_c|$ samples, which they are willing to use to train the AD model, but not willing to transmit over the medium. Each client's equipment is subjected to a degradation process denoted as $\mathbf{D}_{c}=\{ D_{c}\}_{t\geq 0}$, which is influenced by the seasonal effects of the environment. This process can lead the equipment to an abnormal state at time $\tau_{c, i}$ when degradation exceeds a certain threshold $d_f$, which may later result in its failure, which we denote as $T^{f}_{c, i}$. This process is different for each client $c$, thus being the main source of heterogeneity among the individual datasets. Based on the proposed notation, for each sample, we can define the time of life as
\begin{equation}\label{eq:tol}
  T^{life}_{c,i}=min(T, T^{f}_{c,i}).%, i=1:|D_c|, \quad c=1,...,|C|
\end{equation}
We denote a particular sample's sensor measurements $j$ of a client $c$ time $t$ as:
\begin{equation}\label{eq:input}
    \mathbf{x}_{c, i}(t) = [x_{c, i, 1}(t), x_{c, i, 2}(t), ..., x_{c, i, K}(t)]^T.%, t=0:T^{life}_{c,i}. 
\end{equation}
The normal/abnormal state of each sensor recording is indicated by a set of binary labels $\{y_{c, i}(t)\}_{t=0}^{T^{life}_{c,i}}$ to register the change point, i.e.,
\begin{equation}\label{eq:labels}
    y_{c, i}(t) = 
    \begin{cases}
      1 & \text{if $t \geq \tau_{c, i}$} \\
      0 & \text{otherwise}
    \end{cases}.       
\end{equation}

Based on (\ref{eq:input}) and (\ref{eq:labels}), we can define the dataset $D_c$ for a client $c$:
\begin{equation}
    D_c = \{(\mathbf{X}_{c,i}, \textbf{Y}_{c,i})\}_{i=1}^{|D_c|}, 
\end{equation}
with
\begin{equation}
    \mathbf{X}_{c, i} = [\mathbf{x}_{c, i}(0), ..., \mathbf{x}_{c, i}(T^{life}_{c,i})]
\end{equation}
\begin{equation}
    \mathbf{Y}_{c, i} = [y_{c, i}(0), ..., y_{c, i}(T^{life}_{c,i})]
\end{equation}

\begin{figure*}[ht]
    \centering
    \subfloat[Expected operating conditions \label{fig:sample_abnorm}]{
        \includegraphics[width=0.50\textwidth]{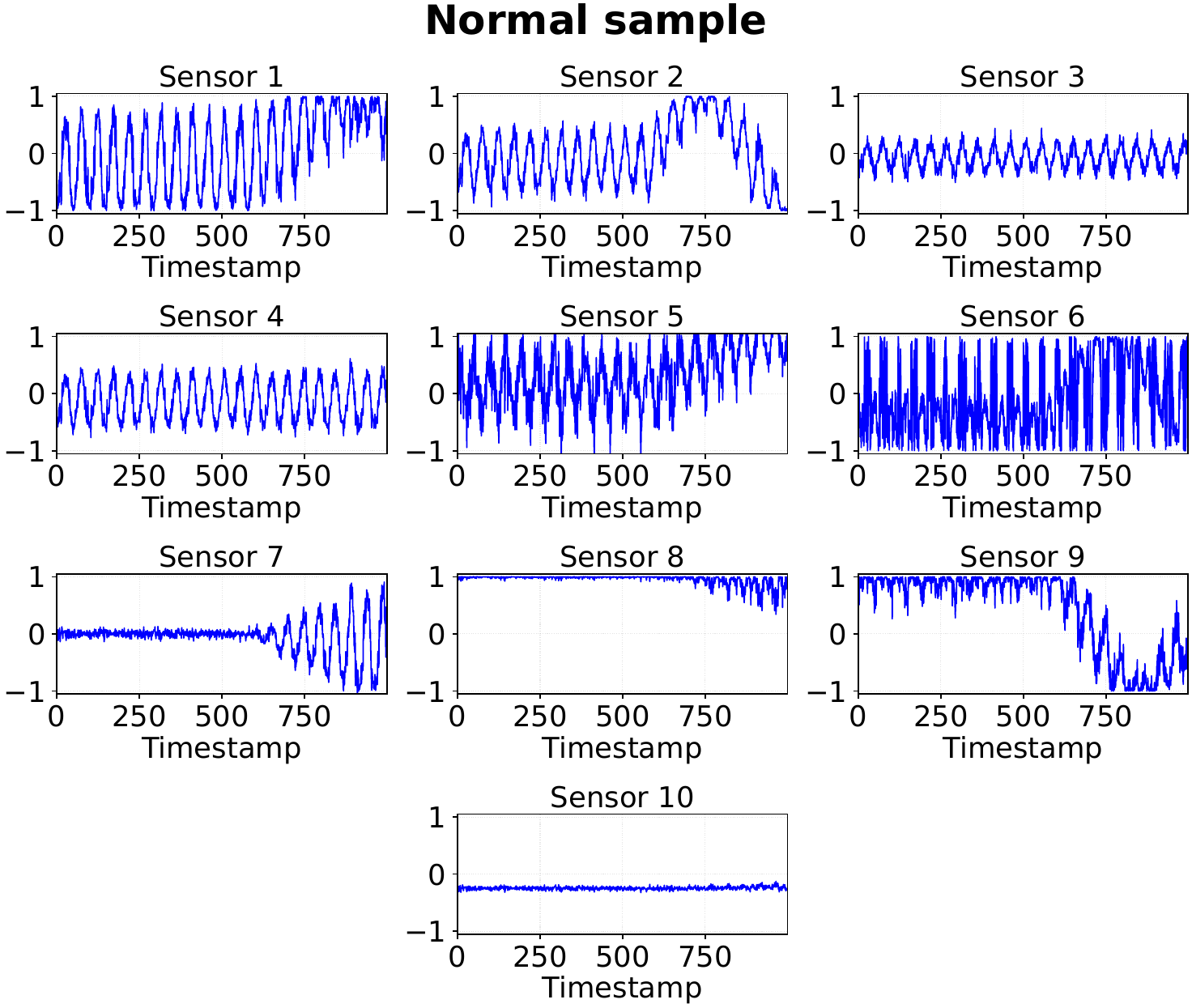}
    }
    \subfloat[Anomalous operating conditions after $\tau$ \label{fig:sample_norm}]{
        \includegraphics[width=0.50\textwidth]{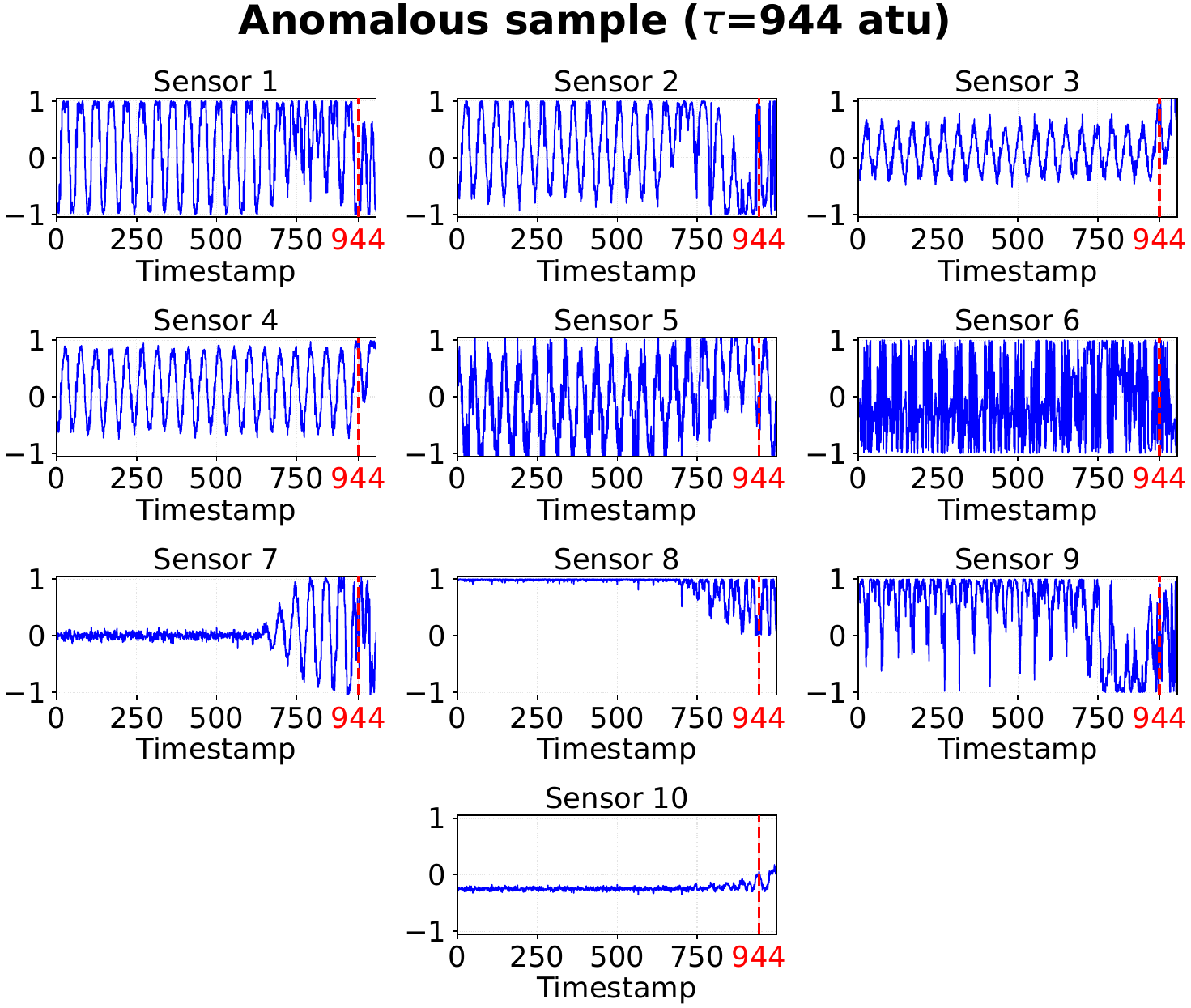}
    }

    \caption{Examples of the Aramis dataset for the selected use-case: a) - sensors operating under normal conditions throughout the simulation time $T$, b) - an example of an anomaly occurring after $\tau=944$ atu (red vertical line).}
    \label{fig:samples}
\end{figure*}

The goal is to collaboratively train either global or personalized generative models for each of the clients, which could accurately estimate the anomaly time $\hat{\tau}_{c.i}$ for the new measurements. The desired model should not only be able to identify the presence of an anomaly, but also estimate the anomaly time as close as possible to the real anomaly time ${\tau}_{c.i}$ to both maximize the operating time and avoid repair costs once the system enters the anomalous state.
\subsection{Generative models}
For VAE, GAN and DDPM in this work, we use approaches and models similar to \cite{VAE_AD_ADOPTED}, \cite{GAN_AD_ADOPTED}, \cite{DDPM_ADOPTED} and \cite{FedSWTSAD} (see Sect. \ref{sec:gener_mod}). The VAE model is constructed of multiple LSTM layers, which we call \textbf{LSTM-VAE}, as defined in \cite{VAE_AD_ADOPTED} and \cite{FEDVAE}. For GAN, we adopt a TAnoGAN-like architecture \cite{GAN_AD_ADOPTED}  and replace the original adversarial objective with the Wasserstein formulation in (\ref{eq:wgan}), together with gradient penalty, to improve training stability \cite{FEDGAN}, \cite{PrivTSAD-FedWGAN}. In addition, we also include \textbf{FedSW-TSAD}, that jointly uses a generative reconstruction component, a discriminator, and a forecasting component, trains using, Sobolev–Wasserstein stabilization. In the federated experiments, the predictor is communicated only in the \emph{Full} federation setup. For the partial federation variants, we follow the same encoder/decoder comparison used for the other models, while the predictor remains local. For DM we use DDPM algorithm, using the Gaussian noise for sample corruption and the model is similar to the classic U-net \cite{UNET}, but with the replacement of 2D convolutions with 1D convolutions, inspired by the U-time \cite{UTIME} architecture and \cite{DDPM_ADOPTED} approach. We call these two approaches \textbf{TAnoWGAN} and \textbf{TAnoDDPM} to differentiate from the original implementations. 
\subsection{Anomaly Detection}
Conventionally, the decision on anomaly time detection relies on a pre-defined anomaly score based on the reconstructed input. For example, for autoencoder- and diffusion-based models usually apply thresholding over the MSE between the input and its reconstruction:
\begin{equation}
    S_{Anom}^{VAE,TAnoDDPM}(\mathbf{x}) = \| \mathbf{x} - \hat{\mathbf{x}} \|_2^2,
\end{equation}
\begin{equation}\label{eq:wgan_loss}
    S_{Anom}^{TAnoWGAN}(\mathbf{x}) = (1-\gamma)\| \mathbf{x} - \hat{\mathbf{x}} \| + \gamma \|h(\mathbf{x})-h(f_\mathbf{\phi}(\mathbf{z}))\|_2^2,
\end{equation}
\begin{equation}
\begin{aligned}
S_{\mathrm{Anom}}^{\mathrm{FedSW\mbox{-}TSAD}}(\mathbf{x})
&= \alpha |\mathbf{x}-\hat{\mathbf{x}}\|_2^2
 + \beta |D_{\bm{\phi}}(\mathbf{x})-\textit{D}_{\bm{\phi}}(\hat{\mathbf{x}})| \\
&\quad + \gamma \|\mathbf{x}^{\mathrm{tar}}-P(\mathbf{x}^{\mathrm{cond}})\|_2^2 \, ,
\end{aligned}
\label{eq:fedsw_score}
\end{equation}
where $P(\cdot)$ the predictor, $\mathbf{x}^{\mathrm{cond}}$ the conditioning input window used for forecasting, and $\mathbf{x}^{\mathrm{tar}}$ the corresponding target future window (see \cite{FedSWTSAD} for details). 

By learning this score for the expected operating conditions, we can select a threshold $\epsilon_c$ that triggers detection when the score exceeds it. We use a client index $c$ to highlight that those thresholds are learned independently during the validation phase (Fig. \ref{fig:system}), which introduces another level of personalization to adapt for local heterogeneity in the data. We use Bayesian optimization \cite{BO} once after training to jointly select the client-specific threshold, a median-filter width that suppresses impulsive noise in the anomaly score, and the minimum number of consecutive positive predictions. On ARAMIS, the objective minimizes the average time offset between the true and predicted anomaly onsets; on SWaT, it maximizes validation F1. This is an offline calibration step and is not part of the federated training loop or online inference.

\subsection{Performance Metric for Change-point Detection}

Most studies assess anomaly-detection performance using standard metrics such as \textit{precision}, \textit{recall} and \textit{F1} score. While these provide a general overview, they often fail to capture the full picture in certain setups. The limitation of the \textit{F1 score} is that it treats each timestamp as an independent and symmetric classification decision. In our setting, however, the problem is sequential, highly imbalanced, and dominated by a single persistent event. Traditional metrics are better suited to spike-anomaly scenarios, in which multiple anomalies may occur during operation and the system does not remain in a faulty state after an anomaly. In our PdM environment, once the system enters an anomalous state, it remains in that state until the end of the simulation or until a random failure occurs. Therefore, metrics such as \textit{F1}, \textit{precision} and \textit{recall} provide only a high-level indication of whether the model detects the anomaly during the simulation, rather than describing how close the predicted anomaly time is to the true one. For completeness, we also report the area under the precision--recall curve (PR-AUC) as a threshold-independent metric commonly used in imbalanced anomaly-detection problems. In our setting, PR-AUC provides a complementary view of how well the anomaly score separates normal from anomalous samples before threshold calibration. However, since PdM decisions depend not only on whether an anomaly is detected but also on \emph{when} it is detected, as discussed previously, PR-AUC alone is not sufficient to assess deployment quality.
 
One option to provide a better general performance of the model is to define a cost for the model being either too sensitive or too conservative. We adopt the loss from \cite{ARAMIS} with the same parameters. The advantage of this loss is asymmetry by design, allowing to set higher costs for the late detection and lower costs for false positives, simulating the real PdM environments. 

The general purpose of the PdM can be described twofold. On one hand, the desired system is expected to operate as long as possible before the expected anomaly or system failure. Thus, a model that would predict the anomaly too soon from the true value will significantly reduce the runtime of the system and increase the cost. Thus, we define the average offset time between the early predicted anomalies and the true anomaly time as:

\begin{equation}
    \Delta t_{FP} = 
    \frac{1}{N_{test}}\sum_{i=1}^{N_{test}}
    \Delta t_{FP}^i 
\end{equation}

\begin{equation}
    \Delta t_{FP}^i =  
    \newline  
    \begin{cases}
      \tau^i - \tau^i_{pred} & \text{if $\tau^i_{pred} \leq \tau^i$} \\
      T_i^{life}-\tau^i_{pred} & \text{if $\tau^i=\varnothing$} \\
      0 & \text{otherwise}
    \end{cases}       
\end{equation}

On the other hand, a conservative system may either predict the anomaly after the system has entered an anomalous state. Given the high cost of repairs, this behavior is even more undesirable than the high sensitivity. We introduce the average offset time between the anomaly and the late predicted anomalies

\begin{equation}
    \Delta t_{FN} = 
    \frac{1}{N_{test}}\sum_{i=1}^{N_{test}}
    \Delta t_{FN}^i 
\end{equation}

\begin{equation}
    \Delta t_{FN}^i =  
    \newline  
    \begin{cases}
      \tau^i_{pred}- \tau^i & \text{if $\tau^i_{pred} \geq \tau^i$} \\
      T_i^{life}-\tau^i & \text{if $\tau^i=\varnothing$}  \\
      0 & \text{otherwise}
    \end{cases}.       
\end{equation}

By analyzing these two metrics, we can get a better understanding of the model behavior, which is more descriptive than the general metrics discussed previously. These metrics provide better guidelines for model selection based on the application, whether the early detection is more desirable than the late detection or the opposite.

\section{Case Studies and Experimental Datasets}\label{sect:case_study}
To validate the proposed federated generative frameworks, we use the ARAMIS dataset \cite{ARAMIS} (Fig.~\ref{fig:samples}) as the primary benchmark and the SWaT dataset \cite{SWAT} as complementary cross-domain validation. ARAMIS remains the main reference for the PdM setting considered in this work, since it matches the persistent-anomaly formulation introduced in Section~\ref{sec: ad_setup_models}, while SWaT is used to assess whether the main trends under partial federation extend to a second industrial TSAD benchmark. For completeness, when we later report complementary results on the SWaT dataset, we restrict the evaluation to \textit{Precision}, \textit{Recall}, \textit{F1}, and \textit{PR-AUC}, as they are sufficient for non-persistent anomalies in TS, while the timing-aware metrics introduced above are specific to the persistent-anomaly PdM formulation considered for ARAMIS. The experiments use a cross-silo setting in which the number of clients is fixed, so that the architecture/federation comparison is controlled and each client retains enough trajectories for meaningful local training.

%\subsection{Dataset and Problem Formulation}
\subsection{ARAMIS Dataset and Problem Formulation}
In the proposed setup, we consider a cross-silo scenario with a small number of reliable institutional clients. Specifically, we set $C=5$ by partitioning the ARAMIS components heterogeneously according to system identifier and environmental degradation process, with one partition per client. Each partition is then split into $|D_c|=30$ complete normal-condition multivariate time-series trajectories for training, $|D^{val}_c|=65$ trajectories for client-specific threshold $\epsilon_c$ calibration, and $|D^{test}_c|=65$ trajectories for testing. Here, $|D_c|$ denotes the number of complete time-series \emph{runs} rather than individual timestamps or windowed instances: each trajectory is monitored by $K=10$ sensors and is subsequently divided into partially overlapping windows of size $20$ with stride $5$, which yields many training windows per trajectory. This regime is chosen to reflect a realistic data-scarce cross-silo PdM setting, in which each site holds only a limited number of historical normal-condition runs for a given asset type and cannot pool raw trajectories across sites. The reported asymmetries should therefore be interpreted as evidence for this low-trajectory industrial setting. The choice of $C=5$ provides interpretable client partitions based on system identifier and degradation process while maintaining adequate per-client trajectory budgets for training, validation, and testing. Broader scaling and heterogeneity analyses are left for future work.

Federated training uses up to $R=30$ communication rounds with $E=4$ local epochs per round. We apply early stopping based on validation performance and retain the best-validation checkpoint for testing, so that reported results correspond to validation-selected models rather than to final-round states. Accordingly, $R=30$ acts as an upper communication budget for the non-convex VAE, GAN, and DDPM objectives rather than as an assumption of monotonic round-wise convergence. The remaining optimization settings are identical across centralized, independent, full-federation, and partial-federation experiments: we use the Adam \cite{ADAM} optimizer with learning rate $10^{-4}$ and batch size $64$. The  validation losses obtained under this budget (without early stopping) are reported in Section~\ref{sec:results}-B.

Post-training, each client uses its local validation set to jointly calibrate the personalized anomaly threshold $\epsilon_c$, the median-filter width used to suppress impulsive score noise, and the minimum-consecutive count. We perform 50 iterations of Bayesian optimization \cite{BO}, minimizing the average anomaly-onset offset for ARAMIS and maximizing validation F1 for SWaT.

Unless otherwise stated, every configuration was repeated over $5$ random seeds controlling model initialization, mini-batch ordering, and client-side data shuffling, and we report mean and standard deviation across seeds for \textit{F1}, \textit{precision}, \textit{recall}, \textit{PR-AUC}, and, on ARAMIS, the asymmetric cost. \textcolor{Violet}{For ARAMIS, the experiment seed also determines the system-to-client assignment.} Independent and federated metrics are first averaged over clients and then aggregated across seeds, so that the encoder/decoder asymmetries discussed in Section~VI can be assessed against seed-level variability while keeping the same optimization, data-processing, and threshold-calibration protocol fixed across federation strategies.

\subsection{SWaT Dataset and Federated Partitioning}
\label{subsec:swat_setup}

To complement the ARAMIS case study with a second industrial benchmark, we evaluate the proposed framework on the Secure Water Treatment (SWaT) dataset \cite{SWAT}. SWaT is a multivariate industrial control system dataset collected from a scaled water-treatment testbed and is widely used for TSAD in cyber-physical systems. Although SWaT is not a PdM with the single persistent changepoint structure as ARAMIS, many attacks affect contiguous temporal intervals rather than isolated point anomalies, making it a meaningful complementary benchmark for the AD setting in this work.

Among the candidate benchmarks, SWaT was selected because it is the closest to the anomaly-detection protocol studied here: it supports unsupervised multivariate TSAD, contains temporally extended anomalous intervals, and allows client-wise validation and threshold selection under a federated partition. By contrast, C-MAPSS/N-CMAPSS \cite{CMAPSS,NCMAPSS} primarily target prognostics and remaining-useful-life estimation, while SMAP/MSL \cite{SMAP_MSL} are relevant AD benchmarks but are less representative of the industrial PdM scenario emphasized in this paper.

To emulate the same cross-silo federation setting consistent with the previous setup, we set $C=5$ clients and partition the SWaT sequence chronologically. The normal-operation portion of the dataset is split into five equal contiguous intervals, and one interval is assigned to each client for local training. The subsequent portion containing anomalous behavior is also split into five equal contiguous intervals aligned with the same clients. Each client-specific interval is then divided into validation and test subsets used for local threshold calibration and final evaluation, respectively. This partition preserves temporal locality within each client and avoids mixing future observations across the training and evaluation stages.

For consistency with the ARAMIS experiments, we retain the same window size, stride, federated protocol, and client-wise threshold selection procedure. However, since SWaT is not evaluated as a PdM changepoint detection task in the same operational sense as ARAMIS, the timing-aware metrics introduced in Section~\ref{sec: ad_setup_models} are not directly comparable. Therefore, for SWaT we report \textit{precision}, \textit{recall}, \textit{F1}, and \textit{PR-AUC}, and use these results as complementary cross-domain validation of the proposed partial-federation strategies.
\vspace{-0.4cm}
\subsection{Generative Models Configurations }
The same model configurations are used across both ARAMIS and SWaT in order to isolate the effect of the federation strategy and dataset characteristics from that of architecture-specific hyperparameter changes. In our experiments, we used the following model architectures: 

\subsubsection{LSTM-VAE}
Following \cite{FEDVAE}, for our experiment, we composed the encoder of two LSTM layers of size 256 and 128 with two linear layers of size 5 to represent the mean and standard deviation of the latent space, and the decoder with two LSTM layers of size 128 and 256, respectively.

\subsubsection{TAnoWGAN}
We stabilize original \cite{GAN_AD_ADOPTED} for FL setup \cite{FEDGAN}, following \cite{PrivTSAD-FedWGAN}. The generator and the critic parts follow the same LSTM-based structure as the VAE, with the latent dimension set to 20. We set the gradient penalty $\lambda$ to 10. For AD, we apply 1500 Adam steps with learning rate $10^{-2}$ to minimize (\ref{eq:wgan_loss}) with respect to $\mathbf{z}$. We set $\gamma$ for the loss to 0.1.

\subsubsection{FedSW-TSAD}
We implement the baseline following the original formulation \cite{FedSWTSAD} for all the parts of the system. We set the hidden dimension of the predictor to 128 for both LSTM layers, and use a two-layer predictor with dropout 0.1. The generator is implemented as a TCN with hidden dimensions $\left[64,128,128\right]$, kernel size 3, and dilations 
$\left[1,2,4\right]$, while the discriminator uses hidden dimensions $\left[128, 128\right]$. We use dropout 0.1 in both the generator and discriminator, $\lambda$ to 10. For anomaly scoring, we set the contribution of each terms as 0.35, 0.15, and 0.5. In order to enable fair comparison, we excluded differential privacy (DP) mechanism from the algorithm, leaving the analysis of the impact on DP for the future work over all the models.

\subsubsection{TAnoDDPM}
As mentioned previously, we follow the architecture proposed in \cite{UTIME}. The encoder consists of three residual blocks (two 1D convolutions with kernel size 3, each followed by normalization and a non-linearity), followed by a down-sampling convolution (kernel size 4). The blocks use 32, 64, and 64 filters. The decoder mirrors this structure: each level applies a transposed 1D convolution for up-sampling, concatenates the result with the corresponding encoder skip connection, and processes the fused features with two residual blocks. The diffusion time embedding size is set to 32. During training, we set $T=1000$ with $\beta_0=10^{-4}$ and $\beta_T=2\cdot10^{-2}$. For anomaly score, we noise a sample to a timestep 250 before denoising.
\section{Experimental Results and Benchmarks} \label{sec:results}
In what follows, we provide a comprehensive analysis of the experimental results. We evaluate the trade-offs between model utility, communication efficiency, and stability across five main scenarios discussed below:

\begin{itemize}
    \item \textbf{Centralized Learning}. Corresponds to a case baseline where we have a single node with all the available training data.
    
    \item \textbf{Independent Training}. In this case, we assume the clients are independent and do not engage in the federation. The performance is the average over the local metrics.
    
    \item \textbf{Full Federation}. A conventional federated learning setup, where we apply Algorithm \ref{alg:fedavg} to the entire model parameters. 

    \item \textbf{Encoder/Discriminator Federation}. In this case, to halve the communication overhead, the clients communicate only the parameters $\bm{\phi}$ corresponding to the \textit{analysis part}, which is an Encoder for VAE and DDPM (of the U-Net model) or a Discriminator for GAN.

    \item \textbf{Decoder/Generator Federation}. Similarly to the previous case, the clients communicate only the \textit{synthesis part} of the model parameters $\bm{\theta}$, which is either a Decoder for VAE and DDPM (of the U-Net model) or a Generator for GAN, to half the communication overhead.
\end{itemize}

Before federated experiments, each model family was tuned centrally under a common $\sim1$M-trainable parameter limit, including learning-rate and architecture/configuration sweeps for TAnoWGAN. FedSW-TSAD instead follows its original configuration as in~\cite{FedSWTSAD} . The selected configurations are fixed across centralized, independent, full-federation, and partial-federation settings, with common federated round budgets, early stopping, and best-validation-checkpoint selection as described in Section~V-A.

\subsection{Results}

% Optional, but helps the tables fit nicely
\newcommand{\ms}[2]{#1{\scriptsize$\pm$}#2}
\setlength{\tabcolsep}{4pt}
\renewcommand{\arraystretch}{1.08}

\begin{table*}[!ht]
\centering
\setlength{\tabcolsep}{4pt}
\renewcommand{\arraystretch}{1.08}
\small

% ---------- (a) ARAMIS ----------
\begin{minipage}[t]{\textwidth}
\centering
\begin{tabular}{llccccc}
\hline
& & \multicolumn{5}{c}{\textbf{Metrics}} \\ \cline{3-7}
\textbf{Case} & \textbf{Model} & \textbf{F1} & \textbf{P} & \textbf{R} & \textbf{PR-AUC} & \textbf{Cost} \\ \hline

\multirow{4}{*}{Centr.}
& LSTM-VAE & \textbf{\ms{0.951}{0.004}} & \textbf{\ms{0.952}{0.005}} & \textbf{\ms{0.950}{0.004}} & \textbf{\ms{0.936}{0.005}} & \textbf{\ms{0.487}{0.009}} \\
& TAnoWGAN & \ms{0.861}{0.014} & \ms{0.840}{0.017} & \ms{0.885}{0.013} & \ms{0.817}{0.016} & \ms{0.583}{0.018} \\
& FedSW-TSAD & \ms{0.889}{0.012} & \ms{0.875}{0.013} & \ms{0.904}{0.011} & \ms{0.853}{0.014} & \ms{0.546}{0.016} \\
& TAnoDDPM & \ms{0.905}{0.009} & \ms{0.896}{0.010} & \ms{0.915}{0.008} & \ms{0.874}{0.010} & \ms{0.523}{0.013} \\ \hline

\multirow{4}{*}{Indep.}
& LSTM-VAE & \ms{0.751}{0.020} & \textbf{\ms{0.827}{0.018}} & \ms{0.688}{0.023} & \textbf{\ms{0.767}{0.019}} & \ms{0.734}{0.026} \\
& TAnoWGAN & \ms{0.698}{0.028} & \ms{0.693}{0.031} & \ms{0.704}{0.027} & \ms{0.676}{0.030} & \ms{0.813}{0.033} \\
& FedSW-TSAD & \ms{0.746}{0.024} & \ms{0.815}{0.021} & \ms{0.689}{0.026} & \ms{0.761}{0.023} & \ms{0.739}{0.028} \\
& TAnoDDPM & \textbf{\ms{0.785}{0.019}} & \ms{0.792}{0.019} & \textbf{\ms{0.779}{0.018}} & \ms{0.758}{0.020} & \textbf{\ms{0.719}{0.024}} \\ \hline
\multirow{4}{*}{Fed. (Full)}
& LSTM-VAE & \textbf{\ms{0.924}{0.006}} & \textbf{\ms{0.927}{0.006}} & \textbf{\ms{0.922}{0.007}} & \textbf{\ms{0.904}{0.007}} & \textbf{\ms{0.507}{0.011}} \\
& TAnoWGAN & \ms{0.828}{0.015} & \ms{0.822}{0.016} & \ms{0.837}{0.014} & \ms{0.790}{0.016} & \ms{0.612}{0.020} \\
& FedSW-TSAD & \ms{0.810}{0.017} & \ms{0.802}{0.018} & \ms{0.818}{0.016} & \ms{0.769}{0.018} & \ms{0.637}{0.022} \\
& TAnoDDPM & \ms{0.874}{0.010} & \ms{0.868}{0.011} & \ms{0.880}{0.010} & \ms{0.840}{0.011} & \ms{0.557}{0.015} \\ \hline

\multirow{4}{*}{Fed. (Enc.)}
& LSTM-VAE & \textbf{\ms{0.909}{0.007}} & \textbf{\ms{0.909}{0.007}} & \textbf{\ms{0.908}{0.008}} & \textbf{\ms{0.884}{0.008}} & \textbf{\ms{0.517}{0.012}} \\
& TAnoWGAN & \ms{0.818}{0.016} & \ms{0.818}{0.017} & \ms{0.819}{0.016} & \ms{0.785}{0.017} & \ms{0.628}{0.021} \\
& FedSW-TSAD & \ms{0.813}{0.018} & \ms{0.813}{0.018} & \ms{0.814}{0.018} & \ms{0.781}{0.019} & \ms{0.633}{0.023} \\
& TAnoDDPM & \ms{0.848}{0.012} & \ms{0.854}{0.012} & \ms{0.844}{0.013} & \ms{0.818}{0.012} & \ms{0.590}{0.017} \\ \hline

\multirow{4}{*}{Fed. (Dec.)}
& LSTM-VAE & \ms{0.869}{0.011} & \ms{0.870}{0.011} & \ms{0.868}{0.012} & \ms{0.839}{0.012} & \ms{0.553}{0.016} \\
& TAnoWGAN & \ms{0.778}{0.019} & \ms{0.776}{0.020} & \ms{0.780}{0.019} & \ms{0.746}{0.021} & \ms{0.683}{0.025} \\
& FedSW-TSAD & \ms{0.749}{0.022} & \ms{0.746}{0.023} & \ms{0.751}{0.022} & \ms{0.718}{0.024} & \ms{0.724}{0.029} \\
& TAnoDDPM & \textbf{\ms{0.884}{0.009}} & \textbf{\ms{0.888}{0.009}} & \textbf{\ms{0.880}{0.010}} & \textbf{\ms{0.858}{0.010}} & \textbf{\ms{0.542}{0.014}} \\ \hline
\end{tabular}
\caption*{(a) Results on ARAMIS dataset}
\end{minipage}

\vspace{0.8em}

% ---------- (b) SWaT ----------
\begin{minipage}[t]{\textwidth}
\centering
\begin{tabular}{llcccc}
\hline
& & \multicolumn{4}{c}{\textbf{Metrics}} \\ \cline{3-6}
\textbf{Case} & \textbf{Model} & \textbf{F1} & \textbf{P} & \textbf{R} & \textbf{PR-AUC} \\ \hline

\multirow{4}{*}{Centr.}
& LSTM-VAE & \textbf{\ms{0.894}{0.007}} & \textbf{\ms{0.911}{0.006}} & \textbf{\ms{0.877}{0.009}} & \textbf{\ms{0.899}{0.007}} \\
& TAnoWGAN & \ms{0.810}{0.016} & \ms{0.823}{0.015} & \ms{0.799}{0.017} & \ms{0.818}{0.016} \\
& FedSW-TSAD & \ms{0.855}{0.013} & \ms{0.867}{0.012} & \ms{0.844}{0.014} & \ms{0.863}{0.013} \\
& TAnoDDPM & \ms{0.883}{0.010} & \ms{0.890}{0.009} & \ms{0.877}{0.011} & \ms{0.888}{0.010} \\ \hline

\multirow{4}{*}{Indep.}
& LSTM-VAE & \textbf{\ms{0.841}{0.015}} & \textbf{\ms{0.860}{0.014}} & \textbf{\ms{0.823}{0.016}} & \textbf{\ms{0.846}{0.015}} \\
& TAnoWGAN & \ms{0.746}{0.021} & \ms{0.758}{0.020} & \ms{0.735}{0.022} & \ms{0.754}{0.021} \\
& FedSW-TSAD & \ms{0.794}{0.018} & \ms{0.808}{0.017} & \ms{0.780}{0.019} & \ms{0.803}{0.018} \\
& TAnoDDPM & \ms{0.830}{0.016} & \ms{0.839}{0.015} & \ms{0.822}{0.017} & \ms{0.835}{0.016} \\ \hline

\multirow{4}{*}{Fed. (Full)}
& LSTM-VAE & \textbf{\ms{0.879}{0.009}} & \textbf{\ms{0.895}{0.008}} & \textbf{\ms{0.864}{0.010}} & \textbf{\ms{0.885}{0.009}} \\
& TAnoWGAN & \ms{0.790}{0.017} & \ms{0.803}{0.016} & \ms{0.779}{0.018} & \ms{0.800}{0.017} \\
& FedSW-TSAD & \ms{0.837}{0.014} & \ms{0.849}{0.013} & \ms{0.827}{0.015} & \ms{0.845}{0.014} \\
& TAnoDDPM & \ms{0.866}{0.011} & \ms{0.873}{0.010} & \ms{0.859}{0.012} & \ms{0.871}{0.011} \\ \hline

\multirow{4}{*}{Fed. (Enc.)}
& LSTM-VAE & \textbf{\ms{0.870}{0.010}} & \textbf{\ms{0.886}{0.009}} & \textbf{\ms{0.855}{0.011}} & \textbf{\ms{0.877}{0.010}} \\
& TAnoWGAN & \ms{0.775}{0.018} & \ms{0.789}{0.017} & \ms{0.762}{0.019} & \ms{0.786}{0.018} \\
& FedSW-TSAD & \ms{0.828}{0.015} & \ms{0.841}{0.014} & \ms{0.816}{0.016} & \ms{0.838}{0.015} \\
& TAnoDDPM & \ms{0.847}{0.012} & \ms{0.856}{0.011} & \ms{0.838}{0.013} & \ms{0.854}{0.012} \\ \hline

\multirow{4}{*}{Fed. (Dec.)}
& LSTM-VAE & \ms{0.853}{0.011} & \ms{0.870}{0.010} & \ms{0.839}{0.012} & \ms{0.860}{0.011} \\
& TAnoWGAN & \ms{0.760}{0.019} & \ms{0.773}{0.018} & \ms{0.747}{0.020} & \ms{0.769}{0.019} \\
& FedSW-TSAD & \ms{0.805}{0.016} & \ms{0.818}{0.015} & \ms{0.793}{0.017} & \ms{0.815}{0.016} \\
& TAnoDDPM & \textbf{\ms{0.871}{0.010}} & \textbf{\ms{0.879}{0.009}} & \textbf{\ms{0.864}{0.011}} & \textbf{\ms{0.877}{0.010}} \\ \hline
\end{tabular}
\caption*{(b) Results on SWaT dataset}
\end{minipage}

\caption{ Performance comparison across centralized, independent, and federated learning settings. Results are reported as mean $\pm$ standard deviation over five random seeds. Table I(a) reports the ARAMIS benchmark, while Table I(b) reports complementary cross-domain validation on SWaT. Independent and federated results are averaged over clients before seed-level aggregation.}
\label{tab:results_main}
\end{table*}

Table~I(a) reports the performance metrics on the ARAMIS dataset, which remains the primary benchmark in this study because it matches the persistent-anomaly PdM setting introduced in Section~\ref{sec: ad_setup_models}. Table~I(b) reports the results on SWaT as complementary cross-domain validation. We discuss ARAMIS first, since only this dataset supports the timing-aware interpretation based on the cost and early/late detection offsets. The SWaT results are then used to assess whether the architectural trends under partial federation remain consistent on a second industrial dataset.

The PR-AUC values in Table~I(a) provide a complementary threshold-independent reference and are broadly consistent with the trends observed for the calibrated metrics: LSTM-VAE and TAnoDDPM remain the strongest architectures, while the evaluated TAnoWGAN configuration is less reliable under the tested protocol. However, in our PdM setting, the practical differences between models are more meaningfully reflected by the time-offset and cost measures, since these capture the operational consequences of early versus late detection. The \textit{Independent Learning} case yields poor performance across all models. The LSTM-VAE, for instance, drops from a \textit{Centralized} \textit{F1-score} of 0.951 to 0.751 when trained locally. The \textit{Full Federation} case significantly bridges the gap: the \textit{Federated} LSTM-VAE achieves an \textit{F1-score} of 0.924, recovering most of the \textit{Centralized} performance. Similarly, TAnoWGAN improves its \textit{F1-score} from 0.698 (\textit{Independent}) to 0.828 (\textit{Full Fed}). FedSW-TSAD provides a useful reference point for modern federated GAN-based TSAD: in the \textit{Centralized} setting, it improves over the plain TAnoWGAN baseline (\textit{F1} = 0.889), confirming the benefit of stronger adversarial stabilization. However, under the tested federated configurations, it does not surpass the best-performing models in our study and remains below LSTM-VAE and TAnoDDPM. This suggests that improved GAN stabilization is beneficial, but does not overturn the broader trend observed here, namely that VAE- and diffusion-based models remain more robust under the considered heterogeneous federated PdM setup.

Analyzing the \textit{partial federation} strategies in Table~I(a) reveals distinct performance outcomes. The VAE achieves higher utility when sharing the \textit{Encoder} (\textit{F1} = 0.909) than when sharing the \textit{Decoder} (\textit{F1} = 0.869). Conversely, the Diffusion Model achieves better performance when sharing the \textit{Decoder} (\textit{F1} = 0.884), slightly above the fully federated model (\textit{F1} = 0.874). In contrast, TAnoWGAN scores decrease in the partial federation settings: both Encoder/Critic and Decoder/Generator sharing yield lower scores (\textit{F1} = 0.818 and 0.778) than full federation (\textit{F1} = 0.828). Regarding the cost of misaligned detection, LSTM-VAE remains the most cost-efficient model in the federated setting (\textit{Cost} = 0.507), closely mirroring the \textit{Centralized} baseline (0.487). TAnoDDPM also maintains stable costs under partial federation, particularly in the Decoder-sharing setup (\textit{Cost} = 0.542), and remains clearly more favorable than TAnoWGAN across the considered federated settings. Several of the partial-vs.-full deltas in Table~I are comparable to the seed-level standard deviation; we therefore read the LSTM-VAE encoder-sharing preference and the TAnoDDPM decoder-sharing preference primarily as cross-dataset qualitative tendencies replicated on ARAMIS and SWaT, rather than as strict numerical orderings within a single benchmark.

\captionsetup[subfigure]{labelformat=empty}
\begin{figure*}[tp]
    \centering
    \subfloat[(a) Centralized Learning \label{fig:time_centr}]{
        \includegraphics[width=0.32\textwidth]{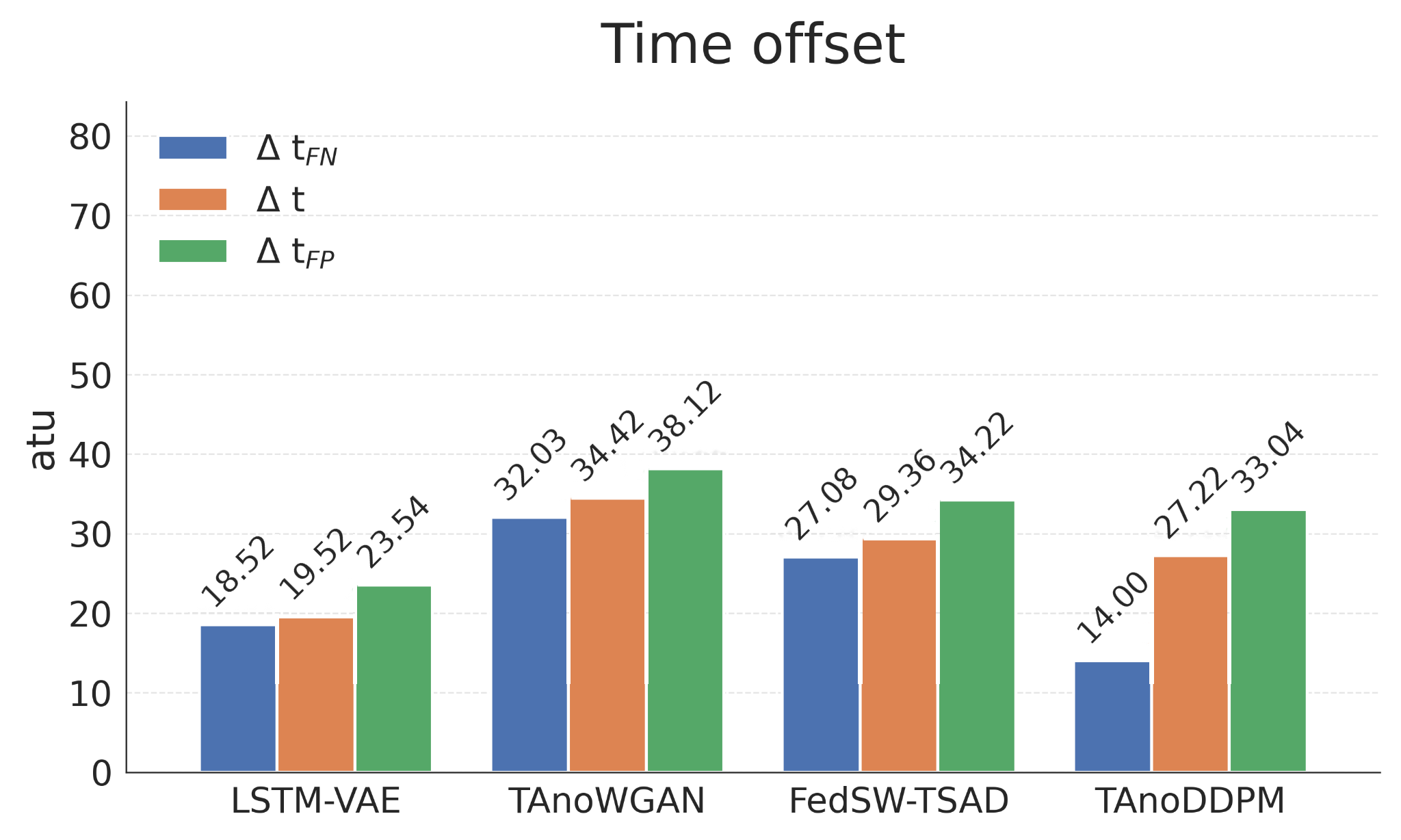}
    }\hspace{-1em} % Adjust horizontal spacing
    \subfloat[(b) Independent Learning \label{fig:time_indep}]{
        \includegraphics[width=0.32\textwidth]{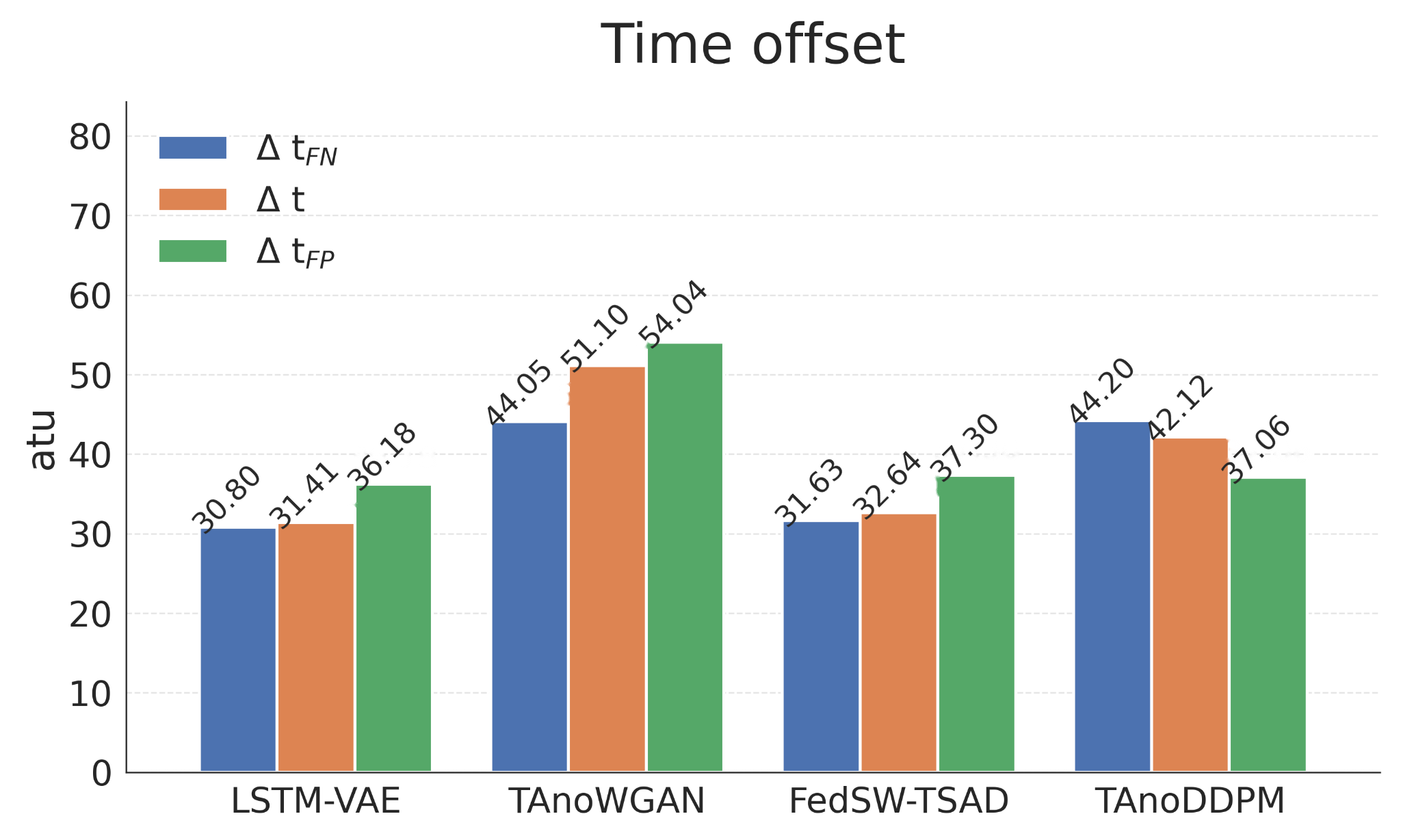}
    }\hspace{-1em} % Adjust horizontal spacing
    \\[-1. em] % Adjust vertical space as needed
    
    %\vspace{-1em}
    \subfloat[(c) Federated Learning (Full) \label{fig:time_fedfull}]{
        \includegraphics[width=0.32\textwidth]{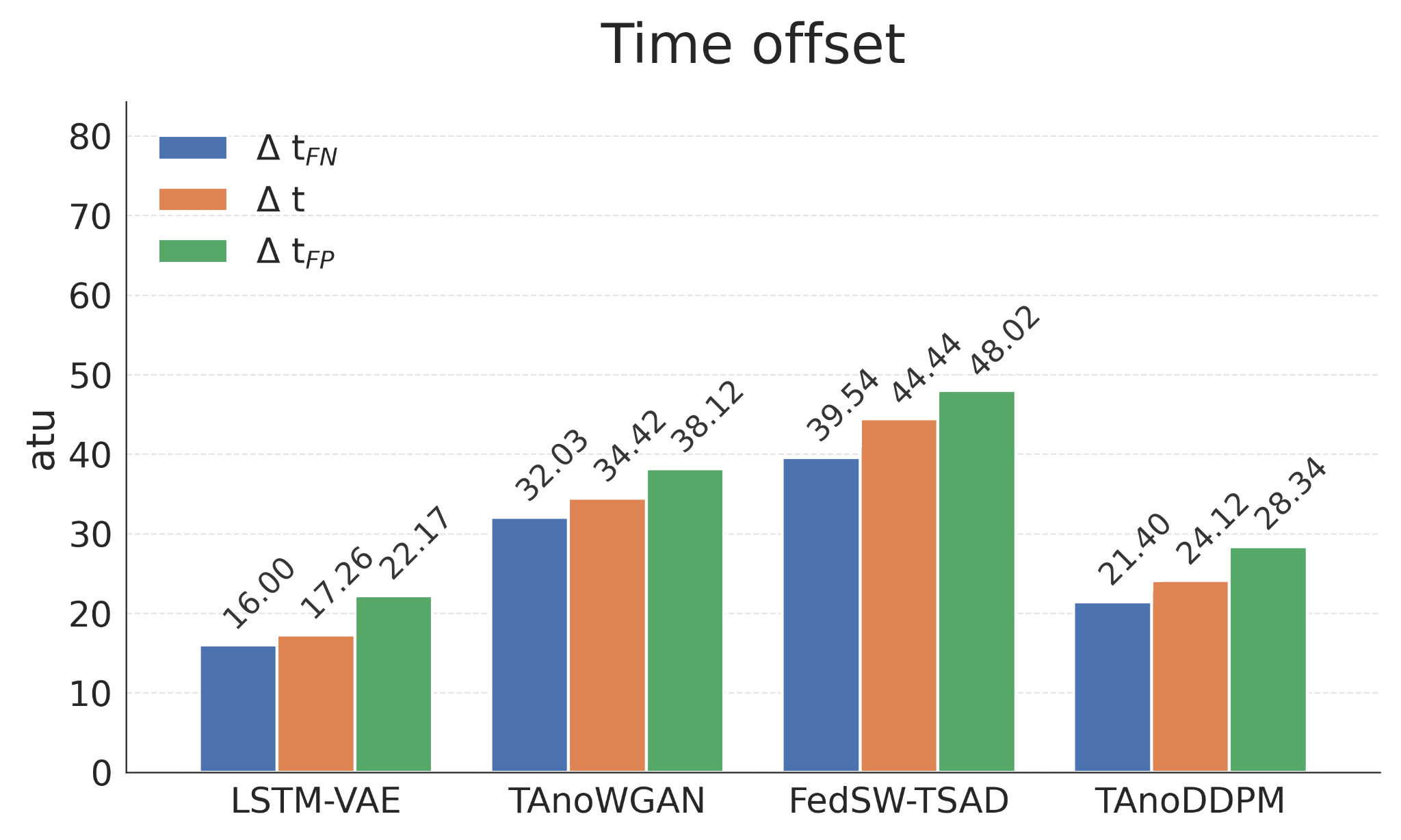}
    }
    \subfloat[(d) Federated Learning (Encoder) \label{fig:time_fedenc}]{
        \includegraphics[width=0.32\textwidth]{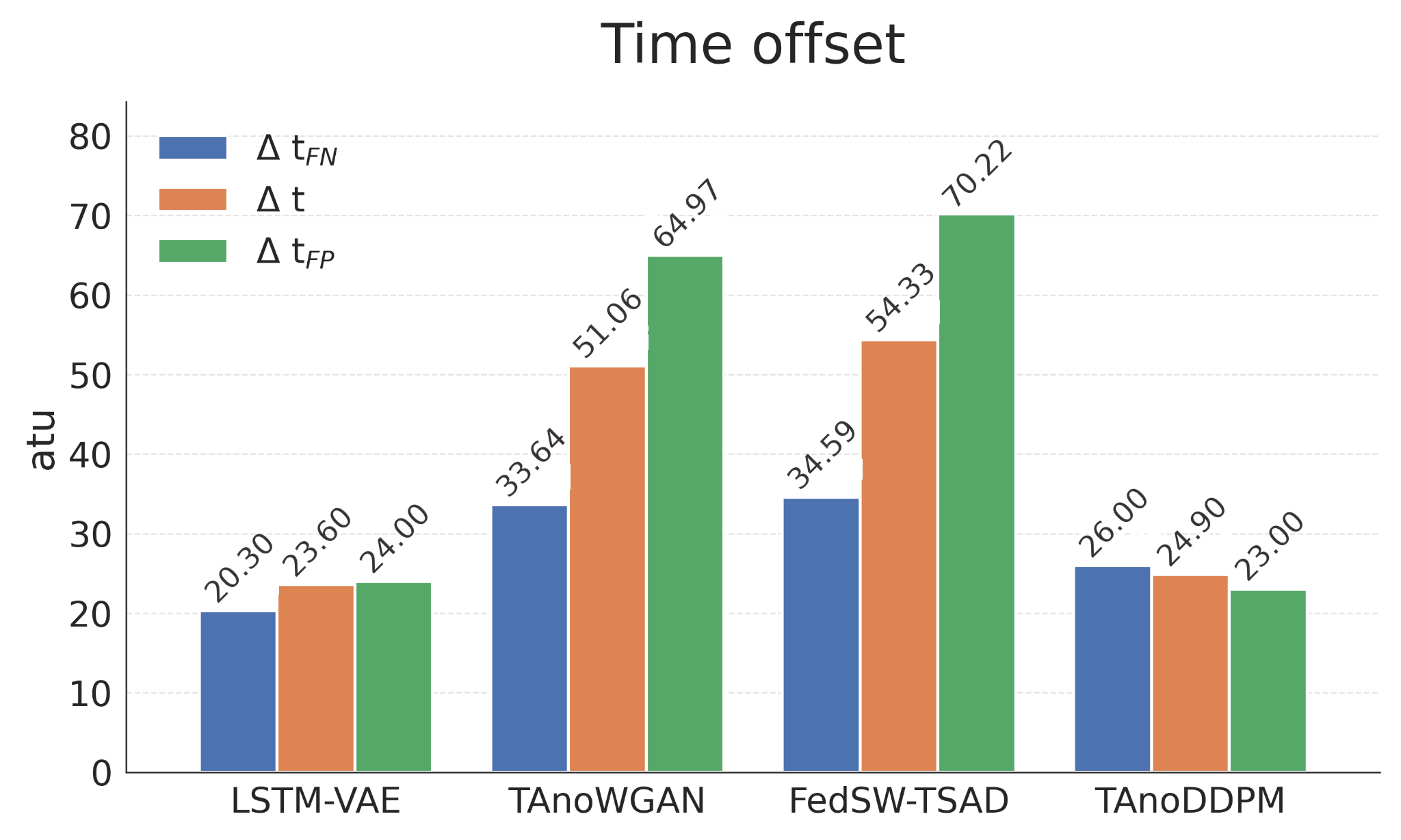}
    } \hspace{-1em} % Adjust horizontal spacing
    \subfloat[(e) Federated Learning (Decoder) \label{fig:time_feddec}]{
        \includegraphics[width=0.32\textwidth]{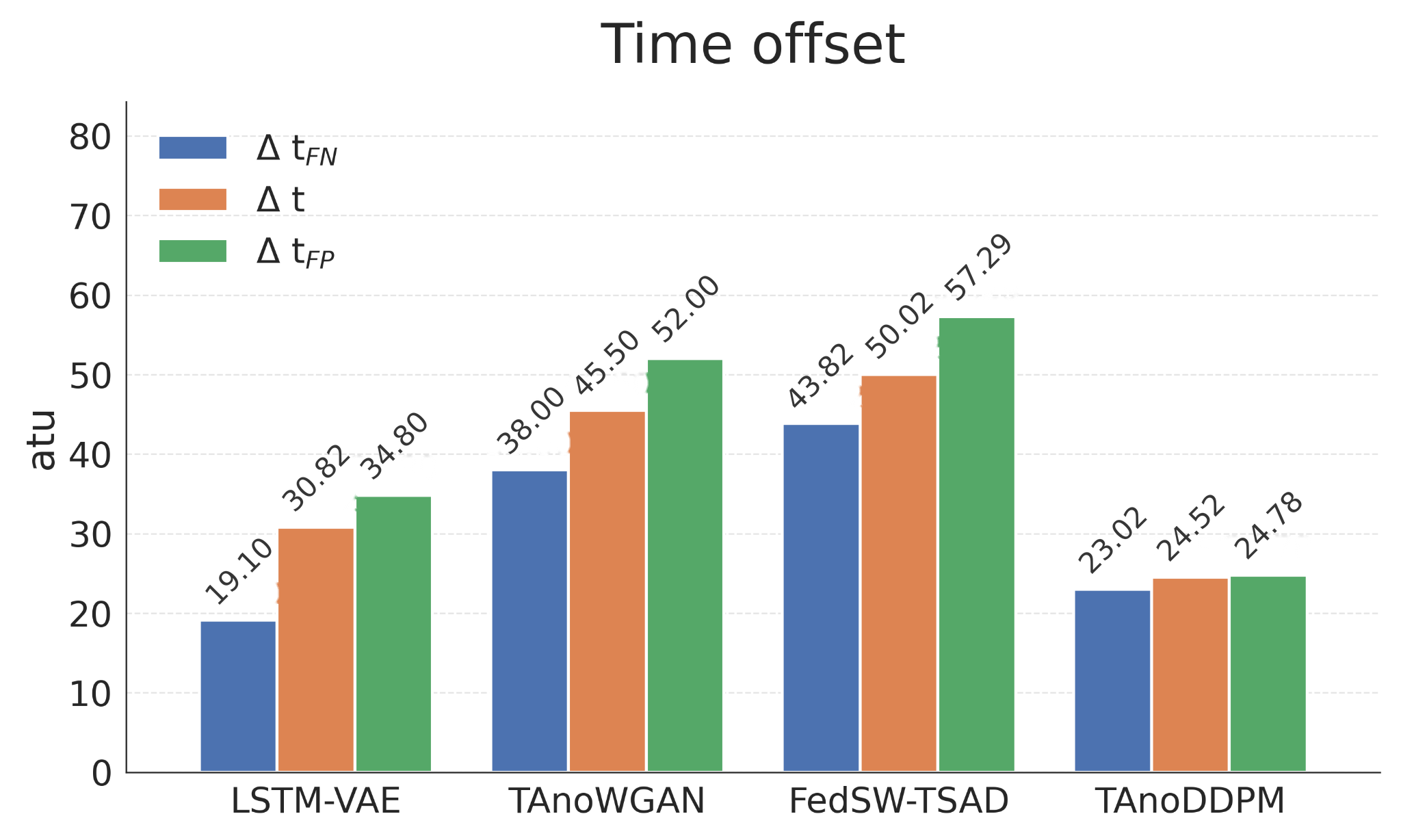}
    } \hspace{-1em} % Adjust horizontal spacing 
    
    \caption{Time offset for different experimental setups: a) - Centralized Learning; b) - Independent Learning; c) Federated Learning with full model transmission; d) - Federated Learning with Encoder/Critic sharing; e) - Federated Learning with Decoder/Generator sharing. \textcolor{Blue}{Blue} color represents the average offset for the late detection; \textcolor{Green}{Green} color represents average early detection offset while \textcolor{Orange}{orange} color shows the average absolute offset.}
    \label{fig:results}
\end{figure*}

Fig. \ref{fig:results} provides a detailed breakdown of the temporal detection performance. It illustrates the average time offset ($\Delta t$) for each model, alongside specific metrics for late detection delays ($\Delta t_{FN}$) and early detection offsets ($\Delta t_{FP}$), across the five experimental configurations. FedSW-TSAD exhibits the largest offsets in all three federated setups (Fig.~\ref{fig:results}c--e). Under Encoder/Critic sharing (Fig.~\ref{fig:results}d), TAnoWGAN has an early-detection offset $\Delta t_{FP}>60$ atu; under Eq.~(20), this quantity denotes an early offset rather than a late-detection delay. LSTM-VAE maintains the lowest offsets under full federation (Fig.~\ref{fig:results}c), comparable to its centralized baseline. The TAnoDDPM profiles under encoder and decoder sharing are broadly comparable: decoder sharing is not uniformly better across the early, late, and absolute offsets.

The complementary results on SWaT in Table~I(b) support the same qualitative interpretation on a second industrial dataset. \textit{Centralized learning} remains the strongest reference, independent training yields a clear degradation, and collaborative training recovers most of the gap. Within partial federation, LSTM-VAE again benefits more from sharing the analysis component than the synthesis component (\textit{F1} = 0.870 vs.\ 0.853), whereas TAnoDDPM remains strongest when sharing the decoder (\textit{F1} = 0.871), slightly above full federation (\textit{F1} = 0.866). TAnoWGAN and FedSW-TSAD both benefit from collaborative training relative to independent learning, but they remain below the best VAE- and DDPM-based configurations. Since SWaT is not evaluated with the timing-aware PdM metrics, these results should be interpreted as complementary evidence that the architectural preferences observed on ARAMIS are not purely dataset-specific.

\subsection{Discussion}
Because the VAE, GAN, and DDPM losses are non-convex, and because clients have heterogeneous time-series data, we do not provide a formal convergence proof. Instead, we assess training stability empirically. To ensure a fair comparison, all FL strategies use the same communication budget, local-training budget, early stopping rule, and best-validation-checkpoint selection. Therefore, the results in Table~\ref{tab:results_main} correspond to validation-selected checkpoints. Full federation also improves final detection utility over independent training across the evaluated models, consistent with clients benefiting from shared information about a broader range of normal operating regimes. The effect of partial versus full federation depends on the model architecture and is discussed next.

\begin{figure*}[!t]
    \centering
    \subfloat[(a) LSTM-VAE \label{fig:conv_vae}]{%
        \includegraphics[width=0.242\textwidth]{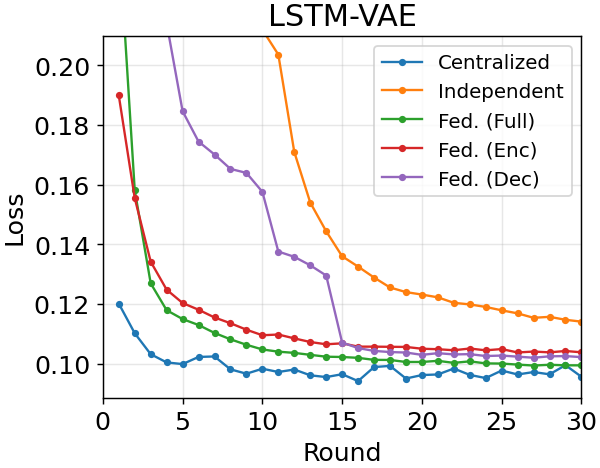}%
    }\hfill%
    \subfloat[(b) TAnoWGAN \label{fig:conv_wgan}]{%
        \includegraphics[width=0.242\textwidth]{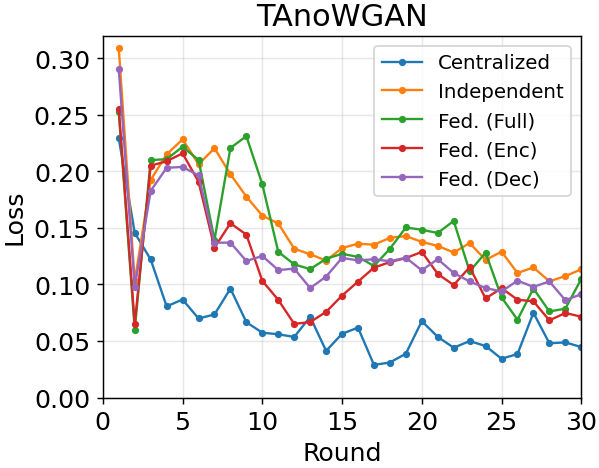}%
    }\hfill%
    \subfloat[(c) TAnoDDPM \label{fig:conv_ddpm}]{%
        \includegraphics[width=0.242\textwidth]{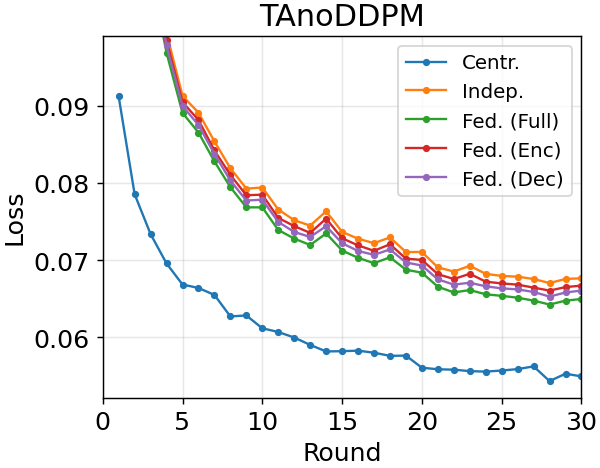}%
    }\hfill%
    \subfloat[(d) FedSW-TSAD \label{fig:conv_fedsw}]{%
        \includegraphics[width=0.242\textwidth]{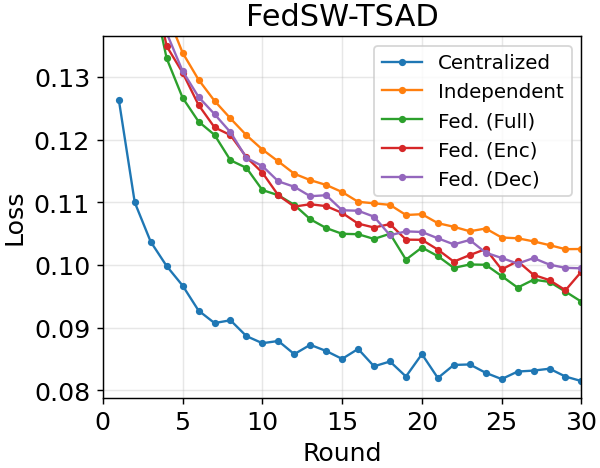}%
    }

    \caption{Per-round validation loss on ARAMIS over the $R=30$ communication rounds for the \textit{Centralized}, \textit{Independent}, \textit{Full fed}, \textit{Fed enc} and \textit{Fed dec} configurations: a) - LSTM-VAE (reconstruction term of the ELBO); b) - TAnoWGAN (Wasserstein-1 estimate in (\ref{eq:wgan})); c) - TAnoDDPM (denoising loss); d) - FedSW-TSAD (predictor loss). Each curve is averaged over clients and the seeds.}
    \label{fig:convergence}
\end{figure*}

Fig.~\ref{fig:convergence} reports the per-round validation loss for the four architectures across the five configurations. For LSTM-VAE and TAnoDDPM, the loss decreases and then plateaus within the round budget, with full and partial federation settling at comparable levels. For FedSW-TSAD the loss is still improving slowly at the end of the budget. TAnoWGAN instead oscillates in every configuration, around a slowly decreasing level and without reaching a flat regime. Combined with Table~II, this indicates that partial federation preserves the stability of full federation at a reduced per-round payload, although it does not improve convergence speed.

The comparison between \textit{Independent Learning} and \textit{Centralized Learning} highlights the critical need for collaborative training in industrial PdM. The poor performance of independent models indicates that local datasets $|D_c|$ are insufficient to capture the full manifold of normal operational conditions. Without exposure to the diverse operating conditions present in the global dataset, local models overfit to narrow patterns and fail to recognize legitimate but unseen normal behaviors, or conversely, fail to reconstruct anomalies that resemble unseen normal data. Collaborative training aggregates this distributed knowledge and improves the final detection behavior relative to independent training across all four model families, including the evaluated adversarial baselines; the adversarial-model conclusions reported below refer specifically to TAnoWGAN and FedSW-TSAD under the configurations and datasets considered here.

A critical insight from this study emerges when analyzing \textit{Partial Federation} strategies, revealing distinct architectural preferences that can guide deployment rules:

\textbf{VAE:} In the VAE framework, the Encoder maps heterogeneous sensor data to a normalized latent space. By sharing the Encoder, clients learn a unified mapping strategy, ensuring that the latent codes are consistent across the federation. The local Decoders can then specialize in reconstructing the specific noise profiles of their local sensors.

\textbf{GAN:} The TAnoWGAN fails to benefit from partial federation. This reinforces the finding that the adversarial equilibrium is fragile; decoupling the generator and critic updates disrupts the training dynamics, leading to a more oscillatory training regime as in Fig.~\ref{fig:convergence}b and to lower final utility. Finally, the timing analysis provides insight into the reliability of these models. The highly dynamic behavior of TAnoWGAN suggests the critic struggles to generalize, causing the model to miss the initial onset of degradation until the signal becomes highly abnormal. The LSTM-VAE's deterministic reconstruction error provides a smooth, reliable signal that crosses the detection threshold promptly after a fault occurs. For TAnoDDPM, decoder sharing maintains a timing profile broadly comparable to encoder sharing while achieving higher F1 and lower cost in Table~I. The comparison with FedSW-TSAD further indicates how well a given adversarial design tolerates heterogeneity and aggregation. In this respect, Sobolev–Wasserstein regularization improves the centralized behavior of adversarial TSAD, yet our results still show a less favorable robustness/communication trade-off than the best VAE- and DDPM-based configurations on this benchmark. All GAN-related observations in this study refer specifically to the evaluated TAnoWGAN and FedSW-TSAD implementations, the ARAMIS and SWaT partitions, and the selected hyperparameter protocol. They should not be interpreted as a general statement about all GAN-based federated TSAD methods.

\textbf{DDPM:} In a significant deviation from the VAE, the Diffusion Model benefits from sharing the \textit{Decoder} (the denoising U-Net's upsampling path). This allows the federation to agree on the fundamental structure of the signal generation/reconstruction, while keeping the Encoder (downsampling/conditioning) local allows for personalized feature extraction. This configuration also yields the lowest cost among the federated DDPM configurations (0.542), making it a highly attractive option for bandwidth-constrained high-fidelity generation.

The SWaT set results reinforce the main conclusions that can be drawn from the ARAMIS set. Although SWaT is operationally different from ARAMIS and is evaluated only with standard detection metrics, the same qualitative tendencies persist: independent local training underperforms collaborative training, VAE benefits more from sharing the analysis component, and DDPM benefits more from sharing the synthesis component. This cross-domain consistency supports the proposed partial-federation taxonomy while keeping ARAMIS as the main benchmark for the timing-aware PdM evaluation.

\begin{table*}[t]
\centering
\caption*{TABLE II: Communication cost per client and round under 32-bit parameter transmission. }
\small
\setlength{\tabcolsep}{4pt}
\begin{tabular}{lccccc}
\hline
Model & Strategy & Shared params. & Download & Upload & Total/round \\
\hline
\multirow{3}{*}{LSTM-VAE}
& Full & 942,956 & 3.77 MB & 3.77 MB & 7.54 MB \\
& Encoder & 473,866 & 1.90 MB & 1.90 MB & 3.79 MB \\
& Decoder & 469,090 & 1.88 MB & 1.88 MB & 3.75 MB \\
\hline
\multirow{3}{*}{TAnoWGAN}
& Full & 961,115 & 3.84 MB & 3.84 MB & 7.69 MB \\
& Encoder & 472,705 & 1.89 MB & 1.89 MB & 3.78 MB \\
& Decoder & 488,410 & 1.95 MB & 1.95 MB & 3.91 MB \\
\hline
\multirow{3}{*}{FedSW-TSAD}
& Full (w. Predictor) & 1,343,855 & 5.38 MB & 5.38 MB & 10.75 MB \\
& Encoder & 131,201   & 0.52 MB & 0.52 MB & 1.05 MB \\
%& Encoder + Predictor & 957,669   & 3.83 MB & 3.83 MB & 7.66 MB \\
& Decoder & 386,186   & 1.54 MB & 1.54 MB & 3.09 MB \\
\hline
\multirow{3}{*}{TAnoDDPM}
& Full & 1,022,922 & 4.09 MB & 4.09 MB & 8.18 MB \\
& Encoder & 544,864 & 2.18 MB & 2.18 MB & 4.36 MB \\
& Decoder & 478,058 & 1.91 MB & 1.91 MB & 3.82 MB \\
\hline
\end{tabular}
\end{table*}

\subsection{Communication and Edge-Device Overhead}
To complement the detection results, we quantify the communication burden of each federation policy during FL process. The client-specific calibration discussed in Sections~IV-C and V-A is performed only once after training on a held-out validation set by jointly optimizing the anomaly threshold, median-filter width, and minimum-consecutive count. It is therefore treated as an offline model-selection step rather than as part of the federated communication rounds or online inference budget. Let $P_{\mathrm{share}}$ denote the number of trainable parameters shared with the server. In each communication round, a participating client downloads $P_{\mathrm{share}}$ parameters and uploads the updated $P_{\mathrm{share}}$ parameters after local training. Hence, the bidirectional number of transmitted parameters per client and round is
\begin{equation}
N_{\mathrm{tx}}^{(r)} = 2P_{\mathrm{share}},
\qquad
V_{\mathrm{tx}}^{(r)} = 8P_{\mathrm{share}} \ \text{bytes},
\label{eq:comm_cost}
\end{equation}
assuming 32-bit floating-point transmission. Over $R$ rounds with $C$ participating clients, the total exchanged volume is
\begin{equation}
V_{\mathrm{tot}} = 8CRP_{\mathrm{share}} \ \text{bytes}.
\label{eq:comm_total}
\end{equation}
Moreover, over a link with bandwidth $B$ bit/s, the communication latency per client and round is
\begin{equation}
T_{\mathrm{comm}}^{(r)} = \frac{64P_{\mathrm{share}}}{B}.
\label{eq:comm_time}
\end{equation}

Table II reports the exact number of shared parameters and the corresponding bidirectional traffic per client and round for the considered federation strategies. As expected, partial federation substantially reduces the communication payload. Compared with full federation, encoder sharing reduces the communicated payload by $49.7\%$, $50.8\%$, $90.2\%$, and $46.7\%$ for LSTM-VAE, TAnoWGAN, FedSW-TSAD, and TAnoDDPM, respectively, while decoder sharing reduces it by $50.3\%$, $49.2\%$, $71.3\%$, and $53.3\%$. The much larger reduction for FedSW-TSAD in the encoder case is due to the fact that the predictor is transmitted only in the full-federation setup, whereas in the partial setups it remains local.

Given that all $C=5$ clients are selected each round, the total footprint over $R=30$ rounds and considering full and partial (encoder/decoder) federation cases is $1.13/0.57/0.56$ GB for LSTM-VAE, $1.15/0.57/0.59$ GB for TAnoWGAN, $1.61/0.16/0.46$ GB for FedSW-TSAD, and $1.23/0.65/0.57$ GB for TAnoDDPM, for \textit{full/encoder/decoder federation}, respectively. For a representative $10$ Mbps edge link, the corresponding communication latency per client and round is $6.03/3.03/3.00$ [sec.] for LSTM-VAE, $6.15/3.03/3.13$ [sec.] for TAnoWGAN, $8.60/0.84/2.47$ [sec.] for FedSW-TSAD, and $6.55/3.49/3.06$ [sec.] for TAnoDDPM. Therefore, FedSW-TSAD has the highest communication cost in the full-federation case, but also benefits the most from partial federation, especially in the encoder-only setup. Since all federation strategies use the same local training budget, the difference in end-to-end training time is due only to the communication term in \eqref{eq:comm_time}.

Partial federation is complementary to standard FL update-compression techniques. Top-$k$ sparsification \cite{TopK_Sparsification}, sparse compression \cite{SparseTernary}, and gradient quantization \cite{FedPAQ} reduce the number of transmitted bits or update coordinates for a selected parameter set, whereas the partial-federation policies studied here structurally reduce the parameter set itself by keeping either the analysis or synthesis module local. The two approaches can be combined: one may first select the shared module according to the proposed analysis/synthesis partition and then apply Top-$k$ sparsification or quantization to the transmitted updates. A systematic evaluation of such combined structural and numerical compression strategies is identified as a concrete future-work direction.

% PART 6 Conclusions
\section{Conclusions} \label{sec: 5_concl}
This work presented a systematic empirical evaluation of split-parameter personalized FL for unsupervised generative TSAD in predictive-maintenance settings. We addressed the dual challenges of data heterogeneity and resource constraints in Industrial IoT by benchmarking VAEs, GANs, and Diffusion Models under partial federation strategies. Our findings lead to three key conclusions. First, collaborative training improves final anomaly-detection utility relative to independent local training across the evaluated models on distributed, non-IID TS data, without sharing raw sensor data. Second, the evaluation across both the primary ARAMIS benchmark and the complementary SWaT dataset confirms that no single architecture is universally optimal, although the main partial-federation trends remain consistent across the two datasets. VAEs yield the highest stability and lowest deployment costs, while DDPMs remain competitive under constrained bandwidth, especially through decoder sharing, which matches full federation at a substantially lower communication cost. In contrast, the evaluated GAN-based baselines, TAnoWGAN and FedSW-TSAD, were less effective in this federated TSAD setting than the VAE- and DDPM-based alternatives under the tested configurations: FedSW-TSAD improved over plain TAnoWGAN in the centralized setting, but neither adversarial model matched the robustness of the VAE- and DDPM-based approaches under federation. This observation should be interpreted as specific to the selected architectures and training setups, rather than as a general conclusion about all GAN-based federated TSAD methods. Finally, we demonstrated that the proposed partial federation approach is not merely a compression technique but a personalization enabler. Specifically, sharing the \textit{Encoder} benefits VAEs, while sharing the \textit{Decoder} benefits DDPMs. Moreover, both LSTM-VAE and TAnoDDPM remain the most reliable and cost-effective choices for general-purpose federated PdM, offering high stability and low computational cost, even with partial federation, thus on average reducing communication overhead by $\approx 50\%$ while achieving a high \textit{F1-score} and small detection offsets.

Because the considered VAE, GAN, and DDPM implementations are highly non-convex neural architectures trained under heterogeneous client distributions, this paper does not claim a formal monotonic-convergence theorem for the resulting federated objectives, nor a closed-form bound on latent collapse or covariate shift. The contribution is instead to provide: i) a structural interpretation of partial federation in terms of shared analysis/synthesis modules, and ii) empirical evidence showing when this partitioning improves stability, utility, and communication efficiency.

Future work will focus on enhancing both the security and efficiency of the proposed federated framework. First, we intend to conduct a rigorous analysis of model convergence under differential privacy (DP) constraints, specifically quantifying the trade-off between strict privacy budgets and the utility of generative models in anomaly detection tasks. Second, we plan to combine the proposed analysis/synthesis partition with optimizer-level personalized FL methods, which we view as complementary rather than alternative to the architectural partitioning studied here. Finally, we plan to evaluate the proposed analysis/synthesis partition under larger client populations and stronger heterogeneity regimes than the cross-silo setting considered here.

% BIBLIOGRAPHY
\bibliographystyle{IEEEtran}
\bibliography{bibliography}

% that's all folks
\end{document}